\documentclass[10pt,twocolumn,letterpaper]{article}

\usepackage[pagenumbers]{wacv} %

\usepackage{mathtools}
\usepackage{multirow}

\definecolor{wacvblue}{rgb}{0.21,0.49,0.74}
\usepackage[pagebackref,breaklinks,colorlinks,allcolors=wacvblue]{hyperref}
\usepackage[accsupp]{axessibility}  %
\def\confName{WACV}
\def\confYear{2027}

\title{DynOSeg: Dynamic Object Segmentation via Motion-Geometry Fusion}

\newcommand{\paperauthors}{Xu Han, Yiduo Wang, Yiran Wang, Viorela Ila\\
Australian Centre for Robotics, School of Aerospace, Mechanical and Mechatronic Engineering, \\
The University of Sydney, NSW, 2006, Australia\\
{\tt\small xhan0329@uni.sydney.edu.au,\{yiduo.wang,yiran.wang,viorela.ila\}@sydney.edu.au}}
\author{\paperauthors}

\usepackage{booktabs}
\usepackage{colortbl}
\usepackage{multirow}
\usepackage{siunitx}
\usepackage{tabularx}

\begin{document}
\maketitle
\begin{abstract}
Moving Object Segmentation (MOS) detects and separates dynamic objects from the static background in a video, 
which plays a fundamental role in dynamic-scene understanding. 
Most MOS methods rely on pure visual motion cues alone, such as optical flow or point tracks, 
which is inherently ambiguous due to the compounded camera and object motions as well as motion-induced appearance changes, e.g. shadows and reflections. 
Naive fusion of additional information such as depth and semantics, however, cannot leverage their complementarity to the fullest extent. 
We therefore propose DynOSeg, a novel MOS approach that takes on a \textbf{two-stage fusion} structure. 
DynOSeg first combines a \textbf{geometry-aware residual map} and depth encoding into a geometric representation of motion, delineating camera motions and object dynamics. We then further incorporate semantic features to refine object understanding. 
We have additionally curated a diverse training corpus including \textbf{CARLA-MOS}, a new synthetic urban-driving dataset for MOS training. 
Our experiments show that DynOSeg consistently outperforms published state-of-the-art MOS methods across a diverse range of datasets. 
Our code and pretrained models can be found at \href{https://github.com/ACFR-RPG/DynOSeg}{https://github.com/ACFR-RPG/DynOSeg}
\end{abstract}
    
\section{Introduction}
\label{sec:intro}

Detecting moving objects with vision alone is a fundamental problem in computer vision and remains an active area of interest for applications such as action recognition~\cite{kong2022human}, Simultaneous Localization And Mapping (SLAM) in dynamic environments~\cite{pi3_zhong2025dynamic, morris2025dynosam}, virtual reality~\cite{wei2025graphavatar} and autonomous driving~\cite{huang2024sgaussian}. 
We address Moving Object Segmentation (MOS), the segmentation of dynamic objects in a video.
Unlike video object segmentation~\cite{video_object_segmentation_survey_yao2020video}, which segments all visible objects, MOS needs to distinguish dynamic objects from the static background. MOS also differs from motion segmentation, which additionally captures background motion such as moving water and shadows.

\begin{figure}[t]
    \centering

    \begin{subfigure}[t]{0.48\linewidth}
        \centering
        \includegraphics[width=\linewidth]{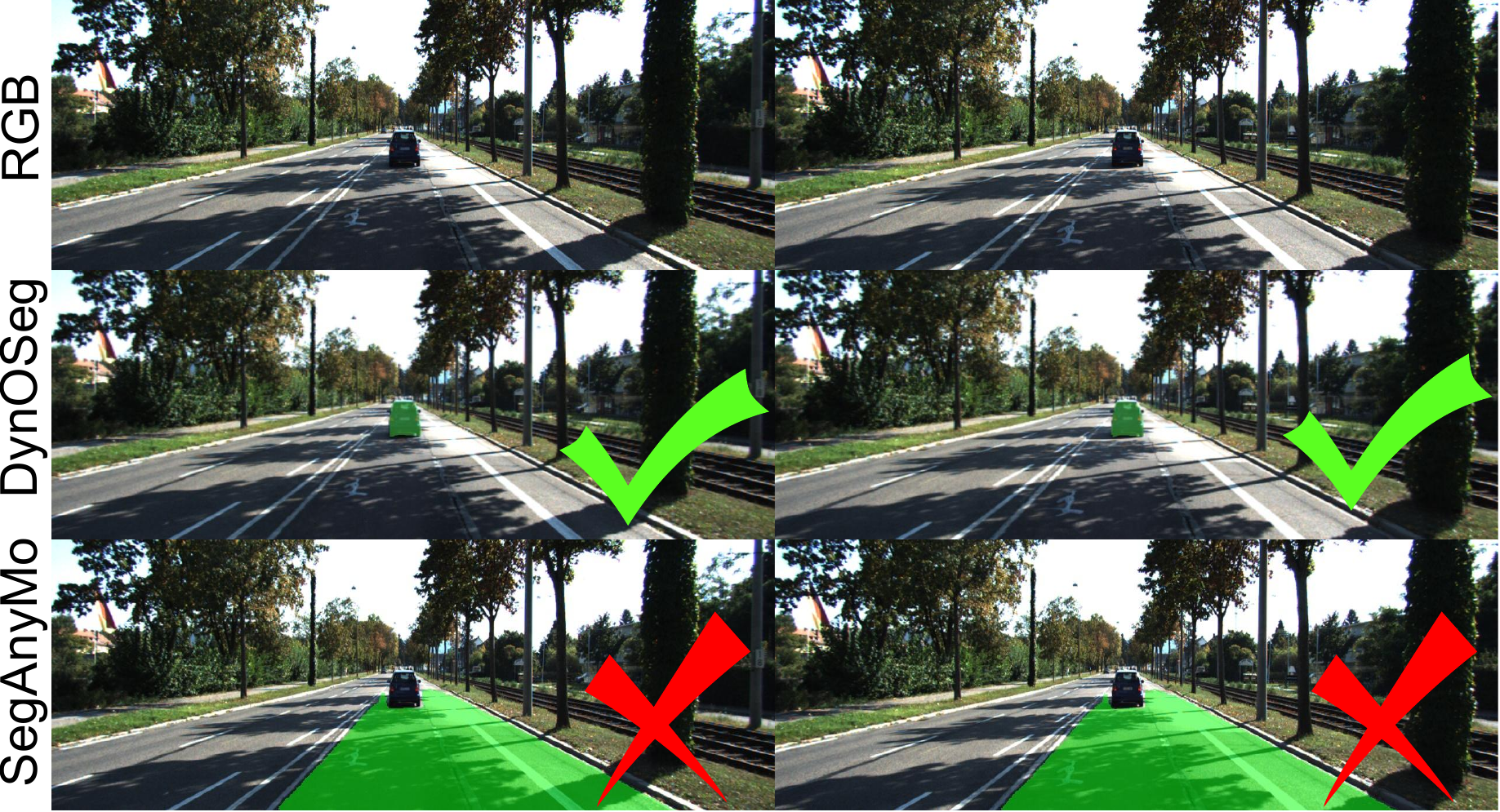}
        \caption{KITTI~\cite{Geiger2013IJRR}.}
        \label{fig:intro-a}
    \end{subfigure}
    \hfill
    \begin{subfigure}[t]{0.48\linewidth}
        \centering
        \includegraphics[width=\linewidth]{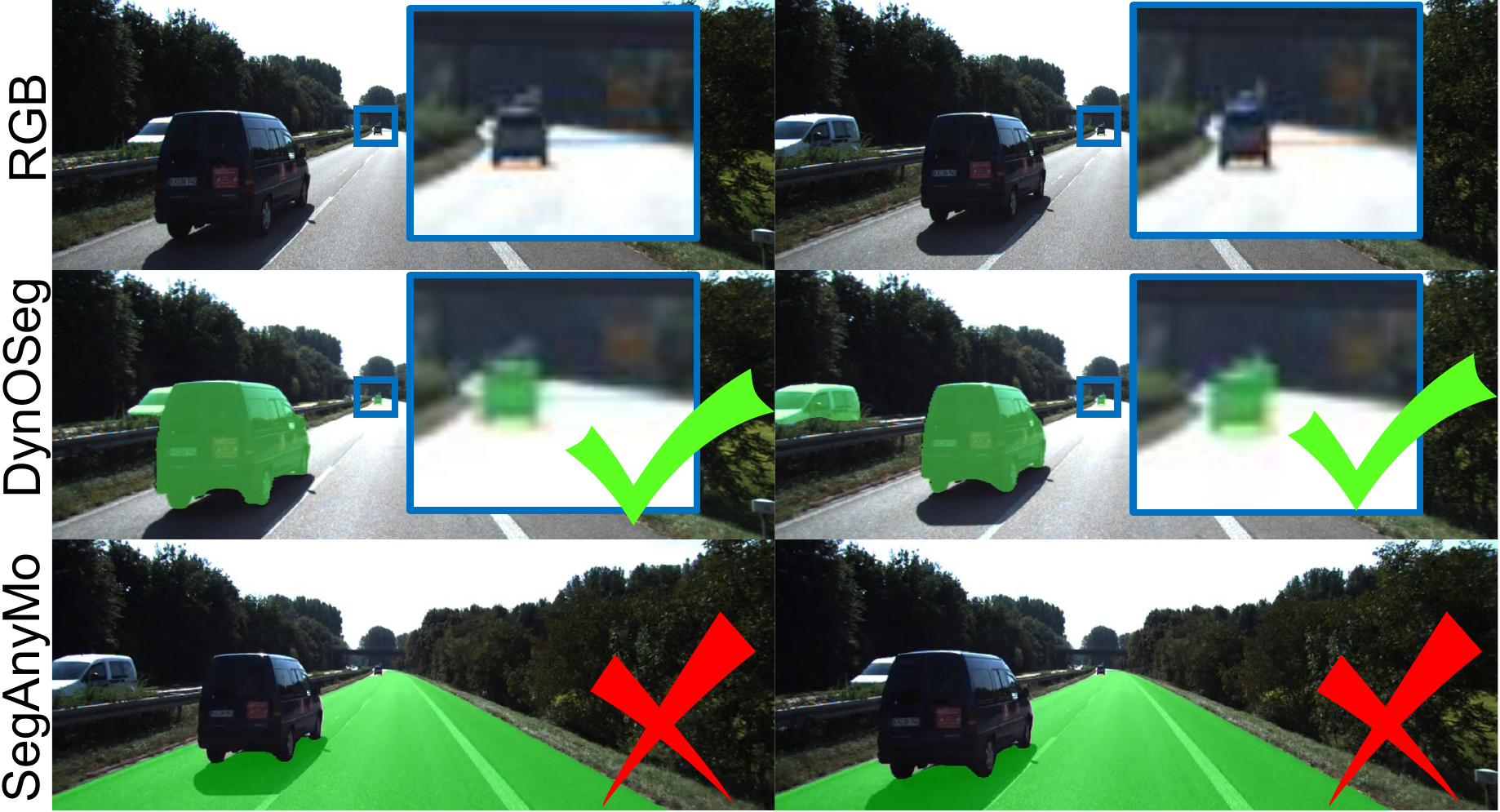}
        \caption{KITTI~\cite{Geiger2013IJRR}.}
        \label{fig:intro-b}
    \end{subfigure}

    \begin{subfigure}[t]{0.48\linewidth}
        \centering
        \includegraphics[width=\linewidth]{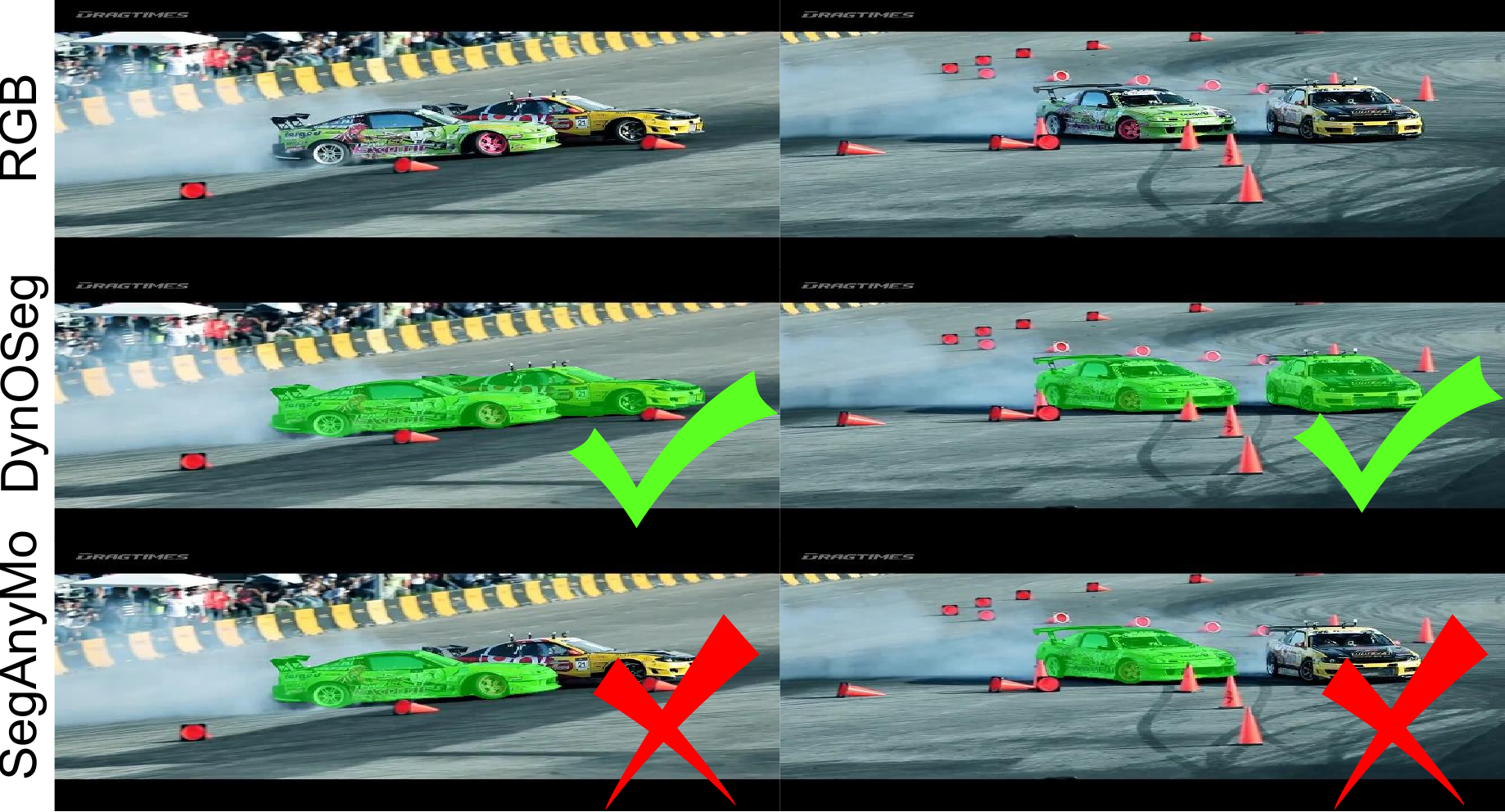}
        \caption{SegTrackV2~\cite{li2013video}.}
        \label{fig:intro-c}
    \end{subfigure}
    \hfill
    \begin{subfigure}[t]{0.48\linewidth}
        \centering
        \includegraphics[width=\linewidth]{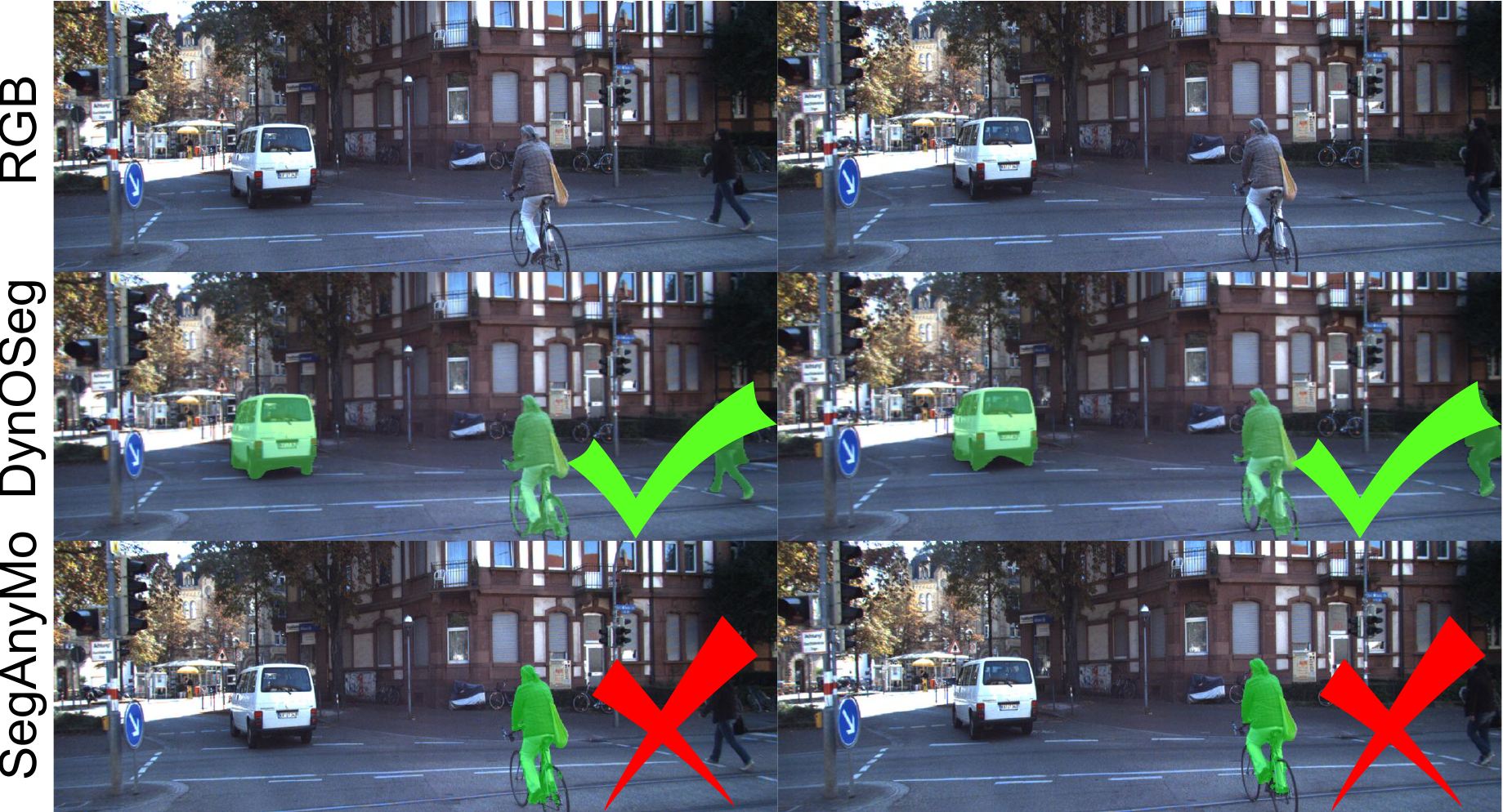}
        \caption{KITTI~\cite{Geiger2013IJRR}.}
        \label{fig:intro-d}
    \end{subfigure}

    \caption{
    \textbf{Challenging scenes.}
    Each panel shows three rows: input RGB, DynOSeg (ours), and
    SegAnyMo~\cite{huang2025segment}, with predicted masks overlaid in the
    second and third rows.
    \textbf{(a--b)} Large camera motion produces substantial background flow;
    in these examples, DynOSeg segments the moving object more accurately than
    SegAnyMo.
    \textbf{(c--d)} In the displayed scenes with multiple moving objects,
    DynOSeg recovers objects that SegAnyMo misses.
    }
    \label{fig:fig1-1}
\end{figure}

Most methods infer motion from optical flow or point tracks~\cite{huang2025segment,huang2024zero,xie24appearrefine,EM}. 
However, these 2D visual motion cues are inherently ambiguous as they reflect the combined effects of camera and object motion, 
making it difficult to separate them using 2D motion cues alone. 
This ambiguity becomes more severe under large camera motion as the static background can exhibit stronger apparent motion than dynamic objects. 
For example, SegAnyMo~\cite{huang2025segment} mistakes the road as dynamic in~\cref{fig:fig1-1} in a highway scene. 
2D appearance cues are also unreliable under illumination changes, shadows, reflections and occlusions. 
These limitations motivate the use of complementary geometry and semantic information. 
Previous works introduce depth and camera geometry to better reason about motion~\cite{Neoral2021bmvc,Homeyer2023iccvw}, 
while recent methods use powerful pretrained DINO/SAM features to provide stronger semantic cues~\cite{huang2025segment, xie2024flowsam}. 
Effective MOS therefore requires combining motion, geometry and semantics to distinguish true object motions and extract accurate object regions.

However, a naive fusion of depth and semantics does not fully exploit unique advantages of these different modalities\cite{rashed2019motion}. 
Depth and visual features provide better geometric and semantic definitions of object boundaries compared to motion-based visual cues. 
For example, the ablation studies of SegAnyMo~\cite{huang2025segment} demonstrate that how and where semantic features are fused with motion cues significantly affects MOS performance.

These observations motivate a modality-aware fusion strategy that preserves the complementary roles of motion, geometry and semantics. Therefore, we propose DynOSeg, a new MOS approach that addresses these challenges. First, DynOSeg uses consistent camera poses and depth together with optical flow to construct a \textit{residual map} that suppresses camera-induced apparent motion. We then employ a two-stage fusion strategy: a Geometry-Aware Spatial Fusion (GSF) block integrates depth and residual encoded features, followed by a Global--Local Semantic-Aware Fusion (GLSF) block that incorporates semantics to refine object regions and boundaries. DynOSeg supports both video and streaming-frame inputs when depth and camera poses are provided by online pipelines.

In summary, our contributions are:
\begin{itemize}
    \item \textbf{Geometry-aware encoding.} We address motion ambiguity by constructing residual map from camera poses, depth, and optical flow, and encoding depth and residual maps with specific encoders to capture geometric cues.

    \item \textbf{Two-stage fusion strategy.} Reliable MOS requires combining camera motion, 3D scene geometry, and semantic cues. Integrating these dependencies, we show that a principled two-stage fusion is able to effectively distinguish moving objects.
    
    \item \textbf{Extensive results.} We conduct extensive experiments and ablation studies to evaluate our method and its key components. We also introduce a new pipeline for generating synthetic MOS datasets with CARLA simulator~\cite{Dosovitskiy17}.
    
\end{itemize}

\section{Related Work}
\label{sec:related}
Visual motion cues, such as optical flow, scene flow, and point tracks, provide the primary evidence for segmenting moving objects in videos~\cite{huang2025segment}. 
Three-dimensional scene geometry and semantic information provide complementary cues for accurate and temporally consistent MOS~\cite{karazija24learning,Homeyer2023iccvw}. 

\textbf{Visual motion cues.}
Optical flow represents the \textit{apparent} motion of image pixels between consecutive frames in a video, 
describing how points move across the image because of camera motion, object motion, or both. 
In scenarios where both camera and scene objects are moving, 
optical flow captures the difference in apparent motions between independently moving objects and the static background, 
which can then be leveraged to detect motion boundaries in the 2D image plane~\cite{Papazoglou2013iccv,Bideau2016eccv} as well as infer underlying 3D motions~\cite{Sekkati2007ras}. 
Some tracking-based MOS methods similarly study the 2D motions between consecutive frames for a sparse collection of points~\cite{Delong2010cvpr,Isack2012ijcv,Barath2019iccv}, 
formulating the MOS problem as model-fitting and solving it via energy minimization. 
Recent progress in machine learning has led to approaches that employ CNNs~\cite{Cao2019cvpr,Faisal2020wacv_eponet,mohamed2021monocular} and transformer-based models~\cite{xie2024flowsam} to infer motion from optical flow, 
or to enhance feature representation and model estimation for tracking-based MOS~\cite{Tokmakov2019ijcv,Zhou2020aaai}.

\textbf{3D geometry consistency.} 
Although learning-based methods have improved MOS accuracy over classical formulations, 
reasoning only over consecutive frames provides limited 3D geometry and long-term motion consistency~\cite{huang2025segment}, 
which can lead to the static background being mistakenly classified as dynamic~\cite{Faisal2020wacv_eponet}. 
To mitigate this limitation in flow-based approaches, 
Wedel~\etal~\cite{wedel2009detection} and Cao~\etal~\cite{Cao2019cvpr} leveraged stereo vision, while Faisal~\etal~\cite{Faisal2020wacv_eponet} incorporated multi-view geometry for 3D understanding.
In comparison, 
multi-frame tracking-based methods study 2D point motion across more than two frames, producing sparse but temporally extended motion representations. 
Longer point tracks enable more coherent clustering~\cite{Ochs2014pami} that is less susceptible to short-term inconsistencies, 
which in turn supports geometric model fitting~\cite{Arrigoni2020eccv,xu2018motion}, subspace fitting~\cite{Elhamifar2013pami,Rao2010pami}, and learned motion segmentation~\cite{huang2024zero,karazija24learning}. 

\textbf{Appearance and semantics.}
Visual motion cues alone may be insufficient to disentangle object dynamics from apparent motion induced by camera movement or depth-dependent motion parallax. 
To address this ambiguity, Mohamed~\etal~\cite{mohamed2021monocular} and Neoral~\etal~\cite{Neoral2021bmvc} integrate semantic information with motion backbones in CNN architectures to better separate independently moving objects from the background.
Similarly, Homeyer~\cite{Homeyer2023iccvw} used a transformer to enable multi-modal fusion between motion and appearance streams for MOS. 
Fusing appearance and semantic information in MOS allows for better distinction between foreground and background~\cite{mohamed2021monocular} and clearer definition of object boundaries~\cite{Neoral2021bmvc}, 
even enabling the segmentation of objects whose motion is partial or non-uniform.

Recent concurrent work shows that 3D geometry and semantic cues provide complementary information for moving-object segmentation~\cite{he2026geomotion,xie2026gmos}. In particular, GeoMotion~\cite{he2026geomotion} combines latent 4D geometry from $\pi^3$ with motion cues, while GMOS~\cite{xie2026gmos} combines geometric features from $\pi^3$ with segmentation features from SAM2 using a single feature-fusion stage. Related studies in dynamic-scene SLAM and reconstruction also demonstrate the value of reliable geometry for reasoning about camera and object motion~\cite{pi3_zhong2025dynamic,morris2025dynosam,Wang2025icra_dynorecon}. For example, $\pi^3_\text{mos}$-SLAM~\cite{pi3_zhong2025dynamic} uses DINOv2~\cite{oquab2023dinov2} features to assist MOS and jointly performs camera pose estimation and scene reconstruction. ViPE~\cite{huang2025vipe}, in contrast, estimates camera motion and scene geometry but does not address MOS. DynOSeg builds on these developments with geometry-aware encoders and two-stage fusion design that integrate camera poses and depth with DINOv2 RGB features, progressively leveraging complementary motion, geometric, and semantic cues for effective moving objects segmentation.

\section{Method}
\label{sec:method}

\begin{figure*}[t]
  \centering
  \includegraphics[width=\linewidth]{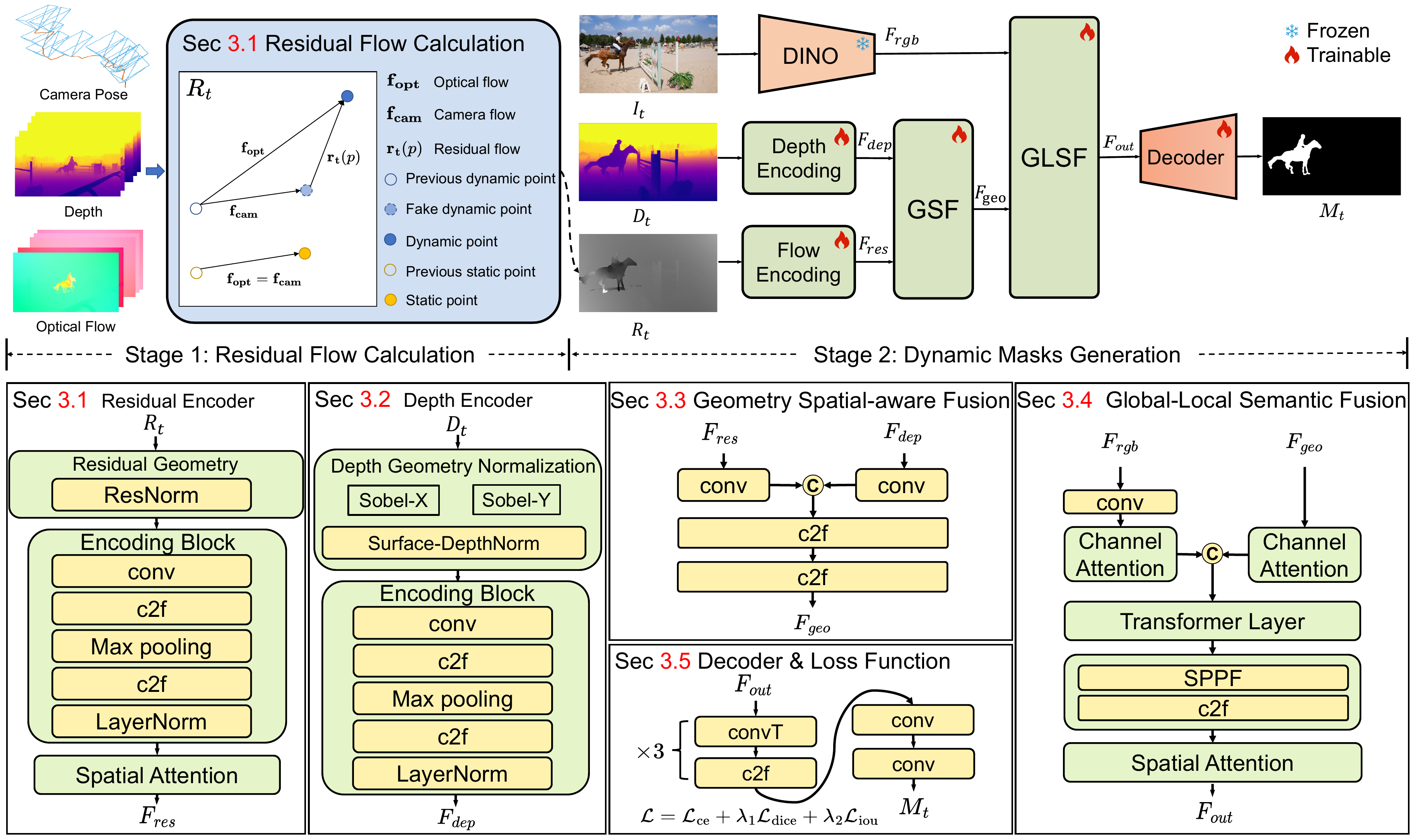} 
  \caption{\textbf{Overview of the proposed DynOSeg framework.} 
  DynOSeg uses depth, camera poses and optical flow to compute a residual map. 
  Depth and residual maps are processed by dedicated encoders, while a frozen DINOv2 encoder~\cite{oquab2023dinov2} extracts high-level semantic features from the RGB image. 
  The resulting multimodal features are fused through Geometry-Aware Spatial Fusion (GSF) and Global--Local Semantic-Aware Fusion (GLSF). Finally, a segmentation decoder processes the fused features to produce moving-object masks.
  }
  \label{fig:model_arch}
\end{figure*}

Applications requiring MOS typically employ perception frontends that already estimate camera poses, dense depth and optical flow. 
DynOSeg is designed to directly leverage these readily available geometric and motion cues. 
In DynOSeg's default setup, 
we use ViPE~\cite{huang2025vipe} for offline camera pose and depth estimation, and RAFT~\cite{teed2020raft} for optical flow, 
but DynOSeg also supports other online inputs such as online a SLAM pipeline, 
as tested and discussed in~\cref{subsec:ablation}.

DynOSeg first constructs a residual map from camera poses, depth and optical flow. 
The depth and residual maps are then processed by dedicated encoding blocks (\cref{sec:depth_encoding,sec:residual}) to produce their corresponding feature representations. 
RGB features are extracted in parallel using DINOv2 encoder~\cite{oquab2023dinov2}, 
which is fully frozen during training and inference. 
Finally, DynOSeg integrates these representations through a \textit{two-stage fusion} strategy that combines residual, depth and RGB features using the proposed GSF and GLSF blocks, described in detail in~\cref{sec:GSF,sec:GLSF}.
The overall architecture is illustrated in~\cref{fig:model_arch}.

\subsection{Residual Map Encoding}
\label{sec:residual}

Residual flow represents the discrepancy between the measured optical flow and the predicted one under the assumption of a fully static environment. 
Existing literature leverages it for 3D scene understanding and separating object motions from camera-induced ones~\cite{taniai2017fast,zou2018dfnet,lee2019learning}. 
DynOSeg also uses this residual flow formulation, and converts residual flow vectors into a scalar residual map as the motion cue input for our model in place of optical flows, 
to exploit information on 3D scene geometry and camera motion for MOS.
Between the current frame $t$ and a previous frame $q$, 
let $\mathbf{T}_{q\leftarrow t}$ define the relative camera transformation that maps camera coordinates from $t$ to $q$. 
Given the depth map of the current frame $D_t$, 
the expected rigid flow for pixel $\mathbf{p}$ under a static-world assumption is:
\begin{equation}
\mathbf{f}_{\mathrm{cam}}(\mathbf{p})
=
\pi\!\left(
\mathbf{T}_{q\leftarrow t}
\left[D_t(\mathbf{p})\,\pi^{-1}(\mathbf{p}), 1\right]^\intercal
\right)-\mathbf{p},
\end{equation}
where $\pi$ and $\pi^{-1}$ denote projection and back-projection functions respectively.
In our current implementation, $q=t-1$ so that DynOSeg studies the motion between consecutive frames, 
inline with optical flow. 
The scalar residual of $\mathbf{p}$ is:
\begin{equation}
R_t(\mathbf{p})
=\left\|
\mathbf{f}_{\mathrm{opt}}(\mathbf{p})
-
\mathbf{f}_{\mathrm{cam}}(\mathbf{p})\right\|_2.
\end{equation}
Thus, static regions are expected to have residuals with small or zero magnitudes, while large residuals indicate motions not caused by camera's own movement.

However, residual flow magnitude varies with object speed, depth and viewing geometry, and may contain spurious responses from depth or pose errors.
The magnitude alone is therefore insufficient for motion detection. 
Hence we encode the residual map rather than using its raw values directly.
Let $\mathbf{R}\in\mathbb{R}^{H\times W}$ denote the residual map.
We first instance-normalize $\mathbf{R}$ to obtain $\hat{\mathbf{R}}$. This tensor is then processed by an encoding block composed of strided convolutions and Cross-Stage Partial (CSP) blocks~\cite{ultralytics}, followed by Layer Normalization, as shown in~\cref{fig:model_arch}. After that,
a spatial attention layer~\cite{woo2018cbam} further emphasizes informative motion regions:
\begin{equation}
\mathbf{F}_{\mathrm{res}}
=
\mathcal{F}_{\mathrm{s}}
\left(
\Phi_{\mathrm{enc}}(\hat{\mathbf{R}})
\right),
\end{equation}
where $\Phi_{\mathrm{enc}}(\cdot)$ denotes the residual encoder and $\mathcal{F}_{\mathrm{s}}(\cdot)$ the spatial attention layer.
The resulting $\mathbf{F}_{\mathrm{res}}$ provides motion features for the subsequent geometry--semantic fusion.

\subsection{Depth Encoding}
\label{sec:depth_encoding}

Depth maps encode important scene geometry but are often corrupted by noise and missing measurements, as well as scale inconsistencies when estimated by per-frame learning-based methods. 
We therefore introduce a Geometry-Aware Depth Encoder to represent local scene and surface geometry.

Given a raw depth map $\mathbf{D} \in \mathbb{R}^{H \times W}$, we first perform normalization using the mean $\mu$ and variance $\sigma^2$~\cite{jun2022depth} over valid depth values ($\mathbf{D} > 0$), yielding a normalized depth map $\hat{\mathbf{D}}$. 
This operation reduces variation in depth ranges across scenes while preserving relative depth discontinuities.
To further encode local geometric structure, we derive an image-space normal representation from $\hat{\mathbf{D}}$ using differentiable Sobel operators~\cite{castleman1979digital}. 
Let $\nabla_x$ and $\nabla_y$ denote the horizontal and vertical depth gradients, respectively. 
The image-space normal map $\mathbf{N}$ is computed as:
\begin{equation}
\begin{aligned}
    \mathbf{N} &=
    \text{Concat}\!\left(
        \frac{-\nabla_x \hat{\mathbf{D}}}{\|\mathbf{v}\|},
        \frac{-\nabla_y \hat{\mathbf{D}}}{\|\mathbf{v}\|},
        \frac{1}{\|\mathbf{v}\|}
    \right) \\
    \|\mathbf{v}\| &=
    \sqrt{(\nabla_x \hat{\mathbf{D}})^2 + (\nabla_y \hat{\mathbf{D}})^2 + 1}.
\end{aligned}
\end{equation}
The normalized depth and image-space normals are concatenated to form a four-channel geometric tensor $\mathbf{I}_{\mathrm{geo}} = [\hat{\mathbf{D}}; \mathbf{N}]$. 
This tensor is then processed by an encoding block, which shares the same architecture as the residual encoder but with independent parameters, to yield depth features $\mathbf{F}_{d}$.

\subsection{Geometry-Aware Spatial Fusion (GSF)}
\label{sec:GSF}

To construct a structural representation, we design a GSF block that fuses depth cues with motion features, as shown in~\cref{fig:model_arch}. 
Since both depth and residual maps encode geometric and motion information about the scene, 
we prioritize their early interaction to form a unified geometric prior, before incorporating  semantic information.

The depth and residual encoders in~\cref{sec:depth_encoding,sec:residual} produce the feature embeddings $\mathbf{F}_{d}$ and $\mathbf{F}_{\mathrm{res}}$, respectively.
We first project them to a shared dimension using $1\times1$ convolutions. 
To fuse these modalities, we concatenate them along the channel dimension and pass them through two Cross-Stage Partial (CSP) blocks~\cite{ultralytics}.
This operation facilitates the fusion of geometry and motion. 
The resulting geometric feature, $\mathbf{F}_\text{geo}$, is formulated as:
\begin{equation}
    \mathbf{F}_\text{geo} = \Phi_\text{csp}\left( \text{Concat}(\mathbf{F}_{d}, \mathbf{F}_{\mathrm{res}}) \right),
\end{equation}
where $\Phi_\text{csp}(\cdot)$ denotes the CSP blocks. 

\subsection{Global--Local Semantic-Aware Fusion (GLSF)}
\label{sec:GLSF}

Building on the geometric features described in~\cref{sec:GSF}, the GLSF block further integrates semantic information and captures cross-modal relationships between semantics and geometry. 

DynOSeg first uses DINOv2~\cite{oquab2023dinov2} to generate RGB features $\mathbf{F}_\text{rgb}$, 
which are projected into the same latent space as $\mathbf{F}_\text{geo}$ using a convolution layer and then refined via a separate Channel Attention gate~\cite{woo2018cbam} to highlight semantic regions of interest. %
In the same block, the geometric features $\mathbf{F}_\text{geo}$ are refined via another Channel Attention gate~\cite{woo2018cbam} to emphasize informative channels that correlate with moving object boundaries.
We then concatenate the refined RGB features with $\mathbf{F}_\text{geo}$. 
This process can be written as:
\begin{equation}
    \hat{\mathbf{F}}_{\text{rgb}} 
    = \mathcal{F}_{\text{CA}}\!\left(\mathcal{F}_{\text{Conv}}(\mathbf{F}_{\text{rgb}})\right), 
    \quad
    \hat{\mathbf{F}}_{\text{geo}} 
    = \mathcal{F}_{\text{CA}}(\mathbf{F}_{\text{geo}}),
    \end{equation}
where $\mathcal{F}_{\text{Conv}}(\cdot)$ denotes a convolutional projection layer and 
$\mathcal{F}_{\text{CA}}(\cdot)$ represents the channel attention operation.

To overcome the limited receptive field of standard convolutions, we introduce a Transformer Layer~\cite{dosovitskiy2020image-vision-t} to model global context across the fused modalities of geometry and semantics.
Following the transformer layer,
we adopt the Spatial Pyramid Pooling-Fast (SPPF) module from YOLOv5~\cite{ultralytics},
combined with a CSP block~\cite{ultralytics} to aggregate multi-scale local context.
Finally, a Spatial Attention mechanism~\cite{woo2018cbam} is integrated to generate the fused features $\mathbf{F}_\text{out}$:
\begin{equation}
    \mathbf{F}_\text{out} = \mathcal{F}_{\text{s}}\left( \Phi_\text{ctx}\left( \mathcal{F}_{\text{t}}(\text{Concat}(\hat{\mathbf{F}}_\text{rgb}, \hat{\mathbf{F}}_\text{geo})) \right) \right),
\end{equation}
where $\mathcal{F}_{\text{t}}(\cdot)$ denotes the Transformer Layer, $\Phi_\text{ctx}(\cdot)$ represents the context aggregation layers (SPPF and CSP), and $\mathcal{F}_{\text{s}}(\cdot)$ denotes the spatial attention.

\subsection{Loss Function}
\label{sec:loss}

A lightweight decoder then uses the fused features $\mathbf{F}_{\text{out}}$ to generate a moving-object mask, as shown in~\cref{fig:model_arch}.
We train the model with a combination of cross-entropy, Dice~\cite{jadon2020survey-dice-loss}, and IoU losses~\cite{yu2016unitbox-iouloss}:
\begin{equation}
\mathcal{L}
=
\mathcal{L}_{\text{ce}}
+
\lambda_1 \mathcal{L}_{\text{dice}}
+
\lambda_2 \mathcal{L}_{\text{iou}},
\end{equation}
where $\mathcal{L}_{\text{ce}}$ supervises static/dynamic classification, while $\mathcal{L}_{\text{dice}}$ and $\mathcal{L}_{\text{iou}}$ are applied to the dynamic-class probability map. We set $\lambda_1=0.5$ and $\lambda_2=0.3$ in all experiments.

\section{Experiments}
\label{sec:exp}
\subsection{Datasets}

DynOSeg is trained on data from six datasets, including our custom-curated CARLA-MOS dataset, which cover a range of scene types and motion patterns. Details are provided in~\cref{tab:data_composition}.
Few datasets are dedicated to MOS, 
as most video segmentation datasets target video object segmentation.
Existing MOS training datasets are predominantly synthetic, which may limit their applicability to real-world scenes. 
For DynOSeg training, we therefore include real-world data drawn from YTVOS~\cite{vos2018}, 
which contributes two subsets,
YTVOS-Moving$^{*}$, an additional manually curated subset containing only sequences with moving target objects, and YTVOS-Moving~\cite{dave2019towards} as shown in~\cref{tab:data_composition}.

\begin{table}[t]
\centering
\caption{Summary of datasets used for training.}
\label{tab:data_composition}
{\small
\begin{tabular}{lccc}
\toprule
\textbf{Dataset} & \textbf{Type} & \textbf{\#Seq.} & \textbf{\#Frames} \\
\midrule
Kubric~\cite{greff2021kubric} & Synthetic & 6,000 & 138k \\
Dynamic Replica~\cite{karaev2023dynamicstereo} & Synthetic & 483 & 145k \\
OCLR-Syn~\cite{OCLR} & Synthetic & 4,600 & 138k \\
CARLA-MOS & Synthetic & 1,000 & 300k \\
YTVOS-Moving~\cite{dave2019towards} & Real-world & 395 & 10.2k \\
YTVOS-Moving$^{*}$ & Real-world & 342 & 9.5k \\
\midrule
\textbf{Total} & -- & \textbf{12,820} & \textbf{740.7k} \\
\bottomrule
\end{tabular}
}
\end{table}

Among the off-the-shelf datasets, 
Kubric~\cite{greff2021kubric} is a synthetic dataset featuring rigid 3D objects with physically plausible dynamics. 
OCLR-Syn~\cite{OCLR} provides synthetic scenes with highly artificial objects and motions.
Both Kubric and OCLR-Syn provide motion masks directly in their ground truth.  
Dynamic Replica~\cite{karaev2023dynamicstereo}, also synthetic, contains only articulated actors such as humans and animals. 
It only provides ground truth camera motion, depth and optical flow, 
hence we created motion masks for DynOSeg training using scene geometry. 
Using the CARLA simulator~\cite{Dosovitskiy17}, we created CARLA-MOS, a synthetic urban-driving dataset curated for MOS, which we will release alongside DynOSeg. The details of the pipeline for constructing with CARLA simulator\cite{Dosovitskiy17} are provided in supplementary materials.

\begin{figure*}[t]
  \centering
  \includegraphics[width=\linewidth]{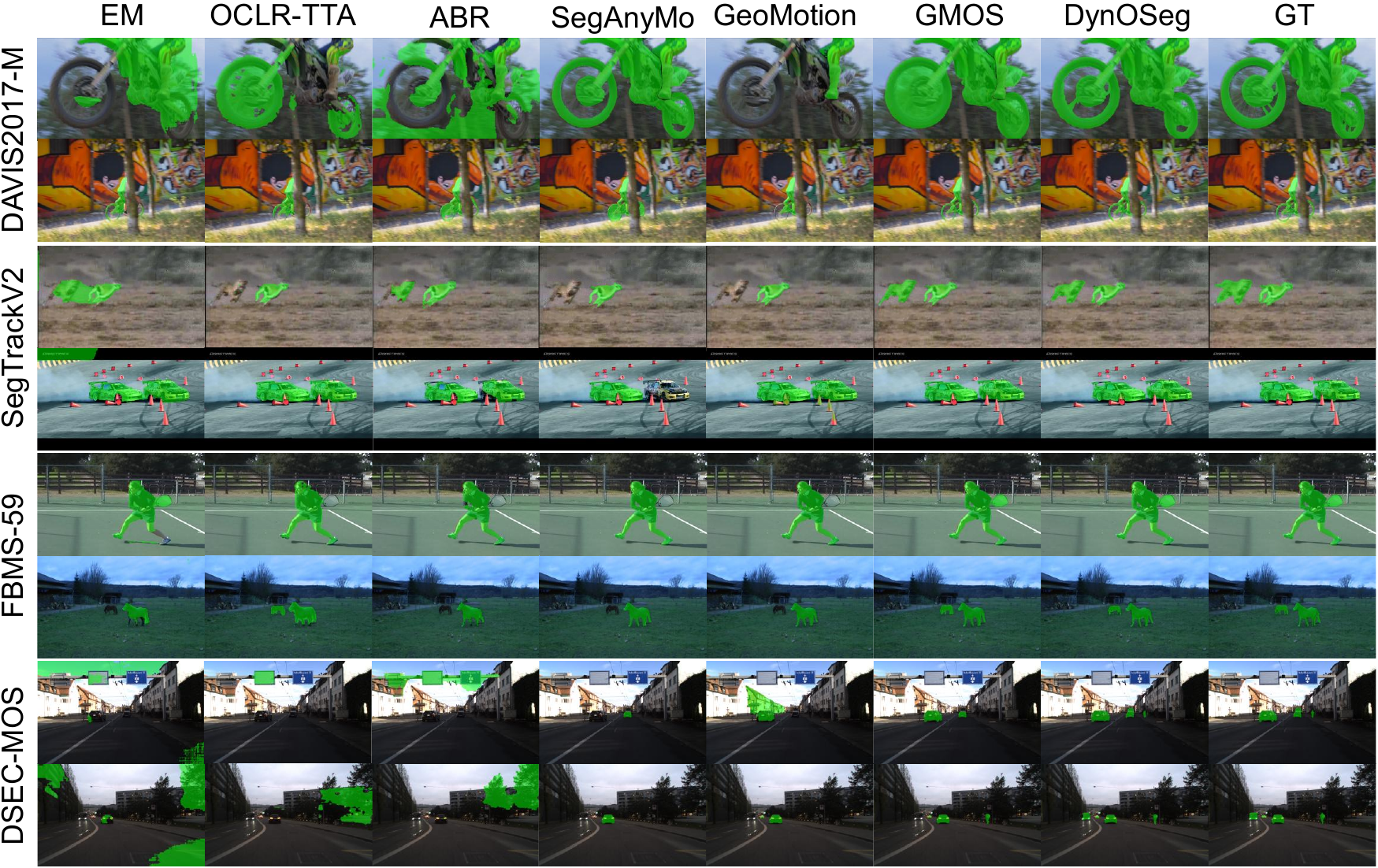} 
  \caption{Qualitative comparisons on selected sequences from the DAVIS2017-M, SegTrackV2, FBMS-59 and DSEC-MOS benchmarks including scenes with multiple moving objects.}
  \label{fig:figure_multiple_moving_objects}
\end{figure*}

\subsection{Experimental Setup and Benchmarks} 
\label{sec:setup}

Unless otherwise stated in the runtime scaling study, DynOSeg uses an input resolution of $518\times518$ on a single NVIDIA RTX 4090.
We use DINOv2~\cite{oquab2023dinov2} (ViT-B/14, frozen throughout training), RAFT~\cite{teed2020raft} (the FlyingThings3D-pretrained checkpoint) and ViPE~\cite{huang2025vipe} (default configuration) as off-the-shelf components without further fine-tuning. 
The whole architecture of DynOSeg contains $5.235$M trainable parameters.

We optimize with Adam (initial learning rate $1\times 10^{-4}$, weight decay $1\times 10^{-5}$) using batch size $4$ for $10$ epochs. 
Training pools the $740.7$k frames across the six training datasets in~\cref{tab:data_composition} and samples per-frame uniformly without per-dataset reweighting, yielding roughly $185$k optimizer steps per epoch and $1.85$M in total.

\textbf{Benchmarks and metrics.}
We evaluate our method across four representative MOS benchmarks with diverse real-world scenarios. 
Specifically, we use the DAVIS2017-M dataset, the standard $22$-sequence moving-object subset of DAVIS~\cite{perazzi2016benchmark} defined by~\cite{dave2019towards}, which contains fast motions and occlusions. 
SegTrackV2~\cite{li2013video} is selected as it provides outdoor videos with strong motion blur and unstructured natural objects. 
FBMS-59~\cite{Ochs2014pami} is a standard moving-object segmentation benchmark containing diverse real-world videos with both camera and object motion.
For autonomous driving scenarios,
we include the official DSEC-MOS test split~\cite{zhou2024event},
featuring densely annotated moving objects such as vehicles and pedestrians. 

\textbf{Local evaluation protocol.}
For all available methods in~\cref{tab:mos}, predicted instances are merged into binary foreground masks and mapped to native ground-truth resolution.
On the official splits, the same evaluation script from SegAnyMo~\cite{huang2025segment} computes region similarity ($\mathcal{J}$), contour similarity ($\mathcal{F}$), and their arithmetic mean $\mathcal{J}\&\mathcal{F}$~\cite{perazzi2016benchmark}.

\begin{table*}[t]
\centering
\scriptsize
\caption{MOS benchmark evaluation. Baseline inference uses official weights and code; all methods use the same SegAnyMo evaluation script~\cite{huang2025segment}. Bold/underline mark best/second-best scores.}
\label{tab:mos}
\setlength{\tabcolsep}{4.3pt}
\renewcommand{\arraystretch}{1.0}
\resizebox{\textwidth}{!}{
\begin{tabular}{l c ccc ccc ccc ccc}
\toprule
\multirow{2}{*}{\textbf{Method}} &
\multirow{2}{*}{\textbf{Year}} &
\multicolumn{3}{c}{\textbf{DAVIS2017-M}} &
\multicolumn{3}{c}{\textbf{SegTrackV2}} &
\multicolumn{3}{c}{\textbf{FBMS-59}} &
\multicolumn{3}{c}{\textbf{DSEC-MOS}} \\
\cmidrule(lr){3-5} \cmidrule(lr){6-8} \cmidrule(lr){9-11} \cmidrule(lr){12-14}
& & $\mathcal{J}\&\mathcal{F}\uparrow$ & $\mathcal{J}\uparrow$ & $\mathcal{F}\uparrow$
& $\mathcal{J}\&\mathcal{F}\uparrow$ & $\mathcal{J}\uparrow$ & $\mathcal{F}\uparrow$
& $\mathcal{J}\&\mathcal{F}\uparrow$ & $\mathcal{J}\uparrow$ & $\mathcal{F}\uparrow$
& $\mathcal{J}\&\mathcal{F}\uparrow$ & $\mathcal{J}\uparrow$ & $\mathcal{F}\uparrow$ \\
\midrule
EM~\cite{EM} & TPAMI'23
& 62.8 & 61.6 & 64.0 & 54.9 & 51.5 & 58.3 & 58.9 & 58.4 & 59.4 & 5.8 & 4.5 & 7.0 \\
OCLR-TTA~\cite{OCLR} & NeurIPS'22
& 69.4 & 69.9 & 69.0 & 63.4 & 61.3 & 65.6 & 63.7 & 63.2 & 64.3 & 6.5 & 4.8 & 8.1 \\
ABR~\cite{xie24appearrefine} & ECCV'24
& 72.3 & 72.7 & 72.0 & 72.4 & 71.1 & 73.7 & 75.0 & 75.7 & 74.3 & 6.3 & 3.5 & 9.2 \\
SegAnyMo~\cite{huang2025segment} & CVPR'25
& 88.5 & 88.0 & 88.9 & 76.3 & 73.5 & 79.1 & 72.6 & 69.4 & 75.9 & 22.6 & 19.9 & 25.2 \\
GeoMotion~\cite{he2026geomotion} & CVPR'26
& 81.2 & 81.1 & 81.3 & 75.3 & 70.9 & 79.8 & 78.8 & 76.7 & 80.9 & 20.5 & 17.3 & 23.8 \\
GMOS~\cite{xie2026gmos} & arXiv'26
& 88.9 & \underline{88.1} & 89.8 & \underline{81.7} & \underline{78.3} & \underline{85.0} & \textbf{92.1} & \textbf{91.0} & \textbf{93.2} & \underline{72.8} & \textbf{68.3} & \underline{77.2} \\
GMOS-S~\cite{xie2026gmos} & arXiv'26
& \underline{89.1} & 87.6 & \textbf{90.7} & 78.0 & 74.6 & 81.4 & \underline{85.1} & \underline{83.7} & 86.5 & 59.3 & 55.3 & 63.2 \\
\midrule
DynOSeg (Ours) & --
& \textbf{90.1} & \textbf{89.6} & \underline{90.6} & \textbf{84.2} & \textbf{81.7} & \textbf{86.6} & 84.6 & 82.6 & \underline{86.6} & \textbf{73.0} & \underline{64.4} & \textbf{81.6} \\
\bottomrule
\end{tabular}
}
\end{table*}

\subsection{Comparison Experiments}

\textbf{Reproduction protocol.} We compare DynOSeg against the available published baselines in~\cref{tab:mos}, including EM~\cite{EM}, OCLR-TTA~\cite{OCLR}, ABR~\cite{xie24appearrefine}, SegAnyMo~\cite{huang2025segment}, and GeoMotion~\cite{he2026geomotion}. For completeness, we additionally report results against the recently released GMOS~\cite{xie2026gmos}, which is currently available only as an arXiv preprint and is therefore not part of the required comparison set. Nevertheless, DynOSeg remains competitive with GMOS across the evaluated benchmarks and outperforms it on several datasets. For all methods, we use the officially released weights and code and evaluate predictions using the same evaluation protocol from SegAnyMo~\cite{huang2025segment}.

\textbf{Standard scenes.}
We define $\Delta$ as the DynOSeg score minus the named comparator's score.
On DAVIS2017-M~\cite{dave2019towards}, DynOSeg leads in $\mathcal{J\&F}$ at $90.1$, ahead of second best GMOS-S at $89.1$ ($\Delta=+1.0$). Its $\mathcal{J}$ of $89.6$ also ranks first, above SegAnyMo at $88.0$ ($\Delta=+1.6$) and GMOS at $88.1$ ($\Delta=+1.5$). Its $\mathcal{F}$ of $90.6$ ranks second to GMOS-S at $90.7$ ($\Delta=-0.1$).

\textbf{Challenging scenes.}
On SegTrackV2~\cite{li2013video}, which includes scenes with multiple dynamic objects, DynOSeg obtains the highest $\mathcal{J\&F}$ score of $84.2$, exceeding GMOS at $81.7$ ($\Delta=+2.5$ points). It also ranks first in $\mathcal{J}$ and $\mathcal{F}$, with gains of $3.4$ and $1.6$ points over GMOS, respectively.
On DSEC-MOS~\cite{zhou2024event}, DynOSeg obtains the highest $\mathcal{J\&F}$ score of $73.0$, exceeding GMOS at $72.8$ ($\Delta=+0.2$ points). DynOSeg has a lower $\mathcal{J}$ score than GMOS ($64.4$ versus $68.3$) but a higher $\mathcal{F}$ score ($81.6$ versus $77.2$). Baselines are evaluated off the shelf from their released checkpoints, whereas DynOSeg is trained on our curated pool. Even without CARLA-MOS, DynOSeg reaches 65.7 on DSEC-MOS, still far above the best published baseline at 22.6, so the margin is not explained by in-domain training data.

The DSEC-MOS results indicate that the improvement in the averaged metric is marginal and is primarily associated with contour accuracy rather than region similarity. 
Figures~\ref{fig:figure_multiple_moving_objects} and~\ref{fig:fig1-1} provide qualitative examples of DynOSeg predictions in scenes with multiple moving objects. In the selected example in~\cref{fig:fig1-1}, SegAnyMo's point-track stage does not fully separate overlapping objects with similar motion, and its subsequent SAM~2~\cite{ravi2024sam2} stage does not recover the occluded object as dynamic.

\subsection{Ablation Study}
\label{subsec:ablation}

In this section, We reported the ablation studies on CARLA-MOS training data, motion representations, input modalities and depth sources, geometry encoders, fusion structures, and different geometry providers.

\textbf{Contribution of CARLA-MOS.}
~\cref{tab:carla_ablation} studies the contribution of CARLA-MOS by removing it from the full training pool while keeping all other datasets unchanged. Removing it reduces $\mathcal{J}\&\mathcal{F}$ by $2.6$ points on DAVIS2017-M and $7.3$ points on DSEC-MOS, showing that CARLA-MOS provides useful training data for motion segmentation.
We attribute this improvement to two main factors. First, CARLA-MOS contains many scenes with multiple moving objects, such as vehicles and pedestrians, which helps the model learn to segment several independently moving objects at the same time. Second, it provides many examples with camera motion along the viewing direction, which is common in real driving scenarios when the camera moves forward or backward with the vehicle. Such motion produces appearance changes that are less common in generic video datasets but are important for autonomous driving. The gain on DSEC-MOS suggests that these driving specific motion patterns help in real road scenes. 

\begin{table}[t]
\centering
\caption{\textbf{CARLA-MOS training-data ablation.}
$Diff.$ denotes the signed change in $\mathcal{J}\&\mathcal{F}$ points relative to the w/ CARLA-MOS setting; negative values indicate a performance drop.}
\label{tab:carla_ablation}
\footnotesize
\setlength{\tabcolsep}{3pt}
\renewcommand{\arraystretch}{1.08}
\begin{tabular*}{\columnwidth}{@{\extracolsep{\fill}}lrrrr@{}}
\toprule
\multirow{2}{*}{Training setting}
& \multicolumn{2}{c}{DAVIS2017-M}
& \multicolumn{2}{c}{DSEC-MOS} \\
\cmidrule(lr){2-3} \cmidrule(lr){4-5}
& $\mathcal{J}\&\mathcal{F}\uparrow$ & $Diff.$
& $\mathcal{J}\&\mathcal{F}\uparrow$ & $Diff.$ \\
\midrule
w/ CARLA-MOS & \textbf{90.1} & $0.0$ & \textbf{73.0} & $0.0$ \\
w/o CARLA-MOS & 87.5 & $-2.6$ & 65.7 & $-7.3$ \\
\bottomrule
\end{tabular*}
\end{table}

\begin{table}[t]
\centering

\caption{\textbf{Motion-representation and input-modality ablation.}
$\dagger$ denotes per-frame depth from Depth Anything V2~\cite{depth_anything_v2},
while $\ddagger$ denotes temporally consistent depth from ViPE~\cite{huang2025vipe}.}
\label{tab:ablation}
\renewcommand{\arraystretch}{1.12}
\footnotesize
\resizebox{\columnwidth}{!}{%
\begin{tabular}{c l cc}
\toprule
\multirow{2}{*}{Exp.} 
& \multirow{2}{*}{Setting}
& DAVIS2017-M
& DSEC-MOS \\
\cmidrule(lr){3-3} \cmidrule(lr){4-4}
& 
& $\mathcal{J}\&\mathcal{F}\uparrow$
& $\mathcal{J}\&\mathcal{F}\uparrow$ \\
\midrule

(A) & Optical Flow + RGB + Depth$^{\ddagger}$ 
& 86.0 & 63.6 \\

\midrule

(B) & Residual Map only
& 55.4 & 61.2 \\

(C) & (B) + Depth$^{\ddagger}$ 
& 79.8 & 64.1 \\

(D) & (B) + RGB 
& 83.5 & 65.6 \\

(E) & (D) + Depth$^{\dagger}$ 
& 78.6 & 63.3 \\

(F) & (D) + Depth$^{\ddagger}$ (DynOSeg)
& \textbf{90.1} & \textbf{73.0} \\

\bottomrule
\end{tabular}%
}
\end{table}

\textbf{Motion representations.}
Rows (A) and (F) of~\cref{tab:ablation} isolate the effect of the motion representation, with RGB and temporally consistent ViPE depth kept unchanged. Replacing optical flow with the residual map improves $\mathcal{J}\&\mathcal{F}$ from $86.0$ to $90.1$ on DAVIS2017-M and from $63.6$ to $73.0$ on DSEC-MOS, giving gains of $4.1$ and $9.4$ points, respectively.
This shows that the residual map provides a cleaner cue for object motion by reducing motion caused by the camera. The larger improvement on DSEC-MOS further highlights its benefit in driving scenes, where strong camera motion is common.

\textbf{Incremental ablation study of DynOSeg.}
Rows (B)--(F) of~\cref{tab:ablation} study the contribution of different input modalities. The residual map alone (B) gives limited performance, while adding either ViPE depth (C) or RGB (D) improves both benchmarks, showing that geometry and appearance provide complementary cues. In contrast, using per-frame Depth Anything V2 depth (E) degrades performance compared with (D). 
We attribute this partly to the scale mismatch between depth and pose translation, as residual-map computation requires depth to be consistent with the metric scale of the pose translation. This change further affects both residual-map construction and the depth input to the depth encoder. Combining the residual map, RGB, and temporally consistent ViPE depth (F) achieves the best results, reaching $90.1/73.0$ $\mathcal{J}\&\mathcal{F}$ on DAVIS2017-M/DSEC-MOS.

\textbf{Encoder ablation.}
Table~\ref{tab:encoder_ablation} compares individual and joint PlainCNN replacements, with inputs, outputs, other modules, and training settings fixed.
PlainCNN uses four Conv--BN--SiLU layers and $0.963$M/$0.961$M depth/residual parameters versus $0.948$M/$0.946$M; details are reported in supplementary materials.
Residual-only and depth-only replacements reduce $\mathcal{J}\&\mathcal{F}$ by 4.1/7.6 and 5.3/8.1 points, compared with DynOSeg on DAVIS2017-M/DSEC-MOS.

\begin{table}[t]
\centering
\caption{\textbf{Encoder ablation.} Listed branches use PlainCNN; others retain their original encoders.}
\label{tab:encoder_ablation}
\footnotesize
\setlength{\tabcolsep}{3pt}
\renewcommand{\arraystretch}{1.08}
\begin{tabular*}{\columnwidth}{@{\extracolsep{\fill}}lcc@{}}
\toprule
\multirow{2}{*}{PlainCNN replacement} & DAVIS2017-M & DSEC-MOS \\
\cmidrule(lr){2-3}
& $\mathcal{J}\&\mathcal{F}\uparrow$ & $\mathcal{J}\&\mathcal{F}\uparrow$ \\
\midrule
Residual encoder only & 86.0 & 65.4 \\
Depth encoder only & 84.8 & 64.9 \\
Both encoders & 85.4 & 65.1 \\
\midrule
None (DynOSeg) & \textbf{90.1} & \textbf{73.0} \\
\bottomrule
\end{tabular*}
\end{table}

~\cref{tab:fusion_ablation} contains five fusion designs. \textit{Concat} directly concatenates the residual, depth, and RGB features. \textit{Only GSF} concatenates the RGB and geometry features from GSF, while \textit{Only GLSF} concatenates the residual and depth features then fuses with RGB features with GLSF. \textit{Swapped order} uses GLSF for geometry fusion and GSF for RGB-geometry fusion. Two-stage fusion performs best, outperforming the strongest alternative, Only GLSF, by $5.6$ points on DAVIS2017-M and $2.4$ points on DSEC-MOS. These results favor the two-stage design under the tested settings.

\begin{table}[t]
\centering
\caption{\textbf{Ablation study on fusion structures.}
All variants use RGB, depth, and residual flow inputs.}
\label{tab:fusion_ablation}
\renewcommand{\arraystretch}{1.12}
\footnotesize
\resizebox{\columnwidth}{!}{%
\begin{tabular}{c l cc}
\toprule
\multirow{2}{*}{Exp.}
& \multirow{2}{*}{Fusion Structure}
& \multicolumn{1}{c}{DAVIS2017-M}
& \multicolumn{1}{c}{DSEC-MOS} \\
\cmidrule(lr){3-3} \cmidrule(lr){4-4}
&
& $\mathcal{J}\&\mathcal{F}\uparrow$
& $\mathcal{J}\&\mathcal{F}\uparrow$ \\
\midrule

(A) & Two-stage
& 90.1 & 73.0 \\

(B) & Concat
& 83.6 & 69.0 \\

(C) & Only GSF
& 82.8 & 69.0 \\

(D) & Only GLSF
& 84.5 & 70.6 \\

(E) & Swapped order
& 83.7 & 69.5 \\

\bottomrule
\end{tabular}
}
\end{table}

\textbf{Plug-and-Play Geometry Sources on KITTI-MOS.}
We evaluate whether DynOSeg can use depth and camera poses from different sources on KITTI-MOS. KITTI-MOS is constructed from KITTI Raw sequences~\cite{Geiger2013IJRR} containing moving objects. The details are provided in supplementary materials.
Table~\ref{tab:kitti_mos_geometry_consistency} compares three geometry settings.
The reference setting combines GT camera poses from KITTI OXTS with
dense stereo depth estimated by FoundationStereo~\cite{wen2025stereo}.
ViPE~\cite{huang2025vipe} estimates both camera poses and depth offline, whereas
DynOSAM~\cite{morris2025dynosam} provides both online. All settings use the same
RGB frames, RAFT optical flow, evaluation masks, and DynOSeg weights. The three geometry sources produce highly similar predictions: $\mathcal{C}_{\mathrm{IoU}}$ ranges from
$84.0$ to $89.0$, with an overall mean pairwise IoU of $87.3$. These results
show that DynOSeg can use reference, offline, or online geometry without
retraining.

\begin{table}[t]
\centering
\caption{
\textbf{Prediction consistency across geometry sources on KITTI-MOS.}
$\mathcal{C}_{\mathrm{IoU}}$ denotes the mean pixel-micro IoU between each configuration and the other two configurations.
}
\label{tab:kitti_mos_geometry_consistency}
\resizebox{\columnwidth}{!}{%
\begin{tabular}{lllc}
\toprule
Pose source & Depth source & Setting
&  $\mathcal{C}_{\mathrm{IoU}}\uparrow$ \\
\midrule
KITTI OXTS (GT) & FoundationStereo & Reference
 & 89.0 \\
ViPE & ViPE & Offline
 & 84.0 \\
DynOSAM & DynOSAM & Online
 & 89.0 \\
\bottomrule
\end{tabular}%
}
\end{table}

\textbf{Online inference.}
The interface uses depth and camera poses from the current and previous frames, together with two frame optical flow, to construct the residual map for the current frame. DynOSeg then takes the current RGB, depth, and residual map as input. By leveraging a causal geometry provider such as DynOSAM or LinStereo~\cite{wang2026linstereo}, our framework can be integrated into an online pipeline. End to end real time performance also depends on the latency of geometry and optical flow.

\textbf{Runtime and Resolution Scaling.}
We further evaluate the computational cost of DynOSeg and its main input components on an RTX 4090 with batch size 1. At $518\times518$, DynOSeg takes $20.13$ ms per frame when depth and residual maps are precomputed; this latency includes DINOv2, data transfer, feature fusion, decoding, and mask post-processing, but excludes RAFT, geometry estimation, residual construction, and disk I/O. As shown in Table~\ref{tab:runtime_analysis}, DynOSeg latency increases from $20.13$ ms at $518\times518$ to $252.69$ ms at $1792\times1008$, while RAFT requires $25.93$--$173.47$ ms and residual construction adds only $0.43$--$1.02$ ms across the tested resolutions. This indicates that residual-map construction adds little overhead compared with optical flow and segmentation. 
Since the GPU times are measured independently, they should not be summed to estimate end-to-end latency and are intended only to characterize the cost of individual components. 
In real applications, DynOSeg works with various sources of optical flow, depth and camera poses.
All tested resolutions are compatible with DINOv2 patch size of $14$, and rectangular inputs use $16{:}9$ aspect ratio. A more detailed analysis is provided in supplementary materials.

\begin{table}[t]
\centering
\caption{\textbf{Runtime scaling on an RTX 4090 (batch size $1$), in
ms/frame.} DynOSeg wall-clock timing includes data transfer and
post-processing; GPU timings are independent compute-only
microbenchmarks and are not additive. RAFT and residual construction
are not included in DynOSeg timing.}
\label{tab:runtime_analysis}
\footnotesize
\resizebox{\columnwidth}{!}{%
\begin{tabular}{lrrrr}
\toprule
Input resolution & \multicolumn{2}{c}{DynOSeg} & RAFT & Residual \\
$(W\times H)$ & Wall-clock & GPU & GPU & GPU \\
\midrule
$518\times518$   & 20.13  & 19.00  & 25.93  & 0.43 \\
$630\times630$   & 32.95  & 32.28  & 31.99  & 0.44 \\
$728\times728$   & 46.93  & 47.15  & 44.89  & 0.46 \\
$1344\times756$  & 102.43 & 106.20 & 84.57  & 0.58 \\
$1792\times1008$ & 252.69 & 270.65 & 173.47 & 1.02 \\
\bottomrule
\end{tabular}%
}
\end{table}

\section{Conclusion and Future Work}
\label{sec:conclusion}
In this paper, we propose DynOSeg, a MOS framework that explicitly models temporally consistent camera motion and depth to segment moving objects in videos. 
DynOSeg addresses the challenge of separating independent object motion from camera-induced motion through a geometry-aware residual map 
and combines camera motion, depth, optical flow, and appearance features through a two-stage fusion strategy. 
In the local benchmark evaluation, DynOSeg ranks first on DAVIS2017-M, SegTrackV2, and DSEC-MOS and third on FBMS-59 in $\mathcal{J\&F}$, with a $0.2$-point margin over GMOS on DSEC-MOS.
The ablation studies support the use of the residual map and temporally consistent depth in the tested DynOSeg configuration. 
One current limitation of DynOSeg is that it does not explicitly distinguish between individual moving-object instances, which we plan to address by incorporating instance-level awareness in future work.

\section*{Acknowledgements}
This research was funded by the Australian Government through the Australian Research Council (ARC) Research Hub for Intelligent Robotic Systems for Real-Time Asset Management (IH210100030).

{
    \small
    \bibliographystyle{ieeenat_fullname}
    \bibliography{main}

\begin{thebibliography}{63}
\providecommand{\natexlab}[1]{#1}
\providecommand{\url}[1]{\texttt{#1}}
\expandafter\ifx\csname urlstyle\endcsname\relax
  \providecommand{\doi}[1]{doi: #1}\else
  \providecommand{\doi}{doi: \begingroup \urlstyle{rm}\Url}\fi

\bibitem[Arrigoni et~al.(2020)Arrigoni, Magri, and Pajdla]{Arrigoni2020eccv}
Federica Arrigoni, Luca Magri, and Tomas Pajdla.
\newblock On the usage of the trifocal tensor in motion segmentation.
\newblock In \emph{Proceedings of the European Conference on Computer Vision (ECCV)}, pages 514--530. Springer International Publishing, 2020.

\bibitem[Baráth and Matas(2019)]{Barath2019iccv}
Dániel Baráth and Jiri Matas.
\newblock Progressive-x: Efficient, anytime, multi-model fitting algorithm.
\newblock In \emph{ICCV}, pages 3779--3787, 2019.

\bibitem[Bideau and Learned-Miller(2016)]{Bideau2016eccv}
Pia Bideau and Erik Learned-Miller.
\newblock It's moving! a probabilistic model for causal motion segmentation in moving camera videos.
\newblock In \emph{Proceedings of the European Conference on Computer Vision (ECCV)}, pages 433--449. Springer International Publishing, 2016.

\bibitem[Cao et~al.(2019)Cao, Kar, Häne, and Malik]{Cao2019cvpr}
Zhe Cao, Abhishek Kar, Christian Häne, and Jitendra Malik.
\newblock Learning independent object motion from unlabelled stereoscopic videos.
\newblock In \emph{CVPR}, pages 5587--5596, 2019.

\bibitem[Carion et~al.(2025)Carion, Gustafson, Hu, Debnath, Hu, Suris, Ryali, Alwala, Khedr, Huang, Lei, Ma, Guo, Kalla, Marks, Greer, Wang, Sun, Rädle, Afouras, Mavroudi, Xu, Wu, Zhou, Momeni, Hazra, Ding, Vaze, Porcher, Li, Li, Kamath, Cheng, Dollár, Ravi, Saenko, Zhang, and Feichtenhofer]{sam3}
Nicolas Carion, Laura Gustafson, Yuan-Ting Hu, Shoubhik Debnath, Ronghang Hu, Didac Suris, Chaitanya Ryali, Kalyan~Vasudev Alwala, Haitham Khedr, Andrew Huang, Jie Lei, Tengyu Ma, Baishan Guo, Arpit Kalla, Markus Marks, Joseph Greer, Meng Wang, Peize Sun, Roman Rädle, Triantafyllos Afouras, Effrosyni Mavroudi, Katherine Xu, Tsung-Han Wu, Yu Zhou, Liliane Momeni, Rishi Hazra, Shuangrui Ding, Sagar Vaze, Francois Porcher, Feng Li, Siyuan Li, Aishwarya Kamath, Ho~Kei Cheng, Piotr Dollár, Nikhila Ravi, Kate Saenko, Pengchuan Zhang, and Christoph Feichtenhofer.
\newblock Sam 3: Segment anything with concepts, 2025.

\bibitem[Castleman(1979)]{castleman1979digital}
Kenneth~R Castleman.
\newblock \emph{Digital image processing}.
\newblock Prentice Hall Professional Technical Reference, 1979.

\bibitem[Dave et~al.(2019)Dave, Tokmakov, and Ramanan]{dave2019towards}
Achal Dave, Pavel Tokmakov, and Deva Ramanan.
\newblock Towards segmenting anything that moves.
\newblock In \emph{2019 IEEE/CVF International Conference on Computer Vision Workshop (ICCVW)}, pages 1493--1502. IEEE, 2019.

\bibitem[Delong et~al.(2010)Delong, Osokin, Isack, and Boykov]{Delong2010cvpr}
Andrew Delong, Anton Osokin, Hossam~N. Isack, and Yuri Boykov.
\newblock Fast approximate energy minimization with label costs.
\newblock In \emph{CVPR}, pages 2173--2180, 2010.

\bibitem[Dosovitskiy et~al.(2017)Dosovitskiy, Ros, Codevilla, Lopez, and Koltun]{Dosovitskiy17}
Alexey Dosovitskiy, German Ros, Felipe Codevilla, Antonio Lopez, and Vladlen Koltun.
\newblock {CARLA}: {An} open urban driving simulator.
\newblock In \emph{Conference on Robot Learning}, pages 1--16, 2017.

\bibitem[Dosovitskiy et~al.(2020)Dosovitskiy, Beyer, Kolesnikov, Weissenborn, Zhai, Unterthiner, Dehghani, Minderer, Heigold, Gelly, et~al.]{dosovitskiy2020image-vision-t}
Alexey Dosovitskiy, Lucas Beyer, Alexander Kolesnikov, Dirk Weissenborn, Xiaohua Zhai, Thomas Unterthiner, Mostafa Dehghani, Matthias Minderer, Georg Heigold, Sylvain Gelly, et~al.
\newblock An image is worth 16x16 words: Transformers for image recognition at scale.
\newblock \emph{arXiv preprint arXiv:2010.11929}, 2020.

\bibitem[Elhamifar and Vidal(2013)]{Elhamifar2013pami}
Ehsan Elhamifar and René Vidal.
\newblock Sparse subspace clustering: Algorithm, theory, and applications.
\newblock \emph{IEEE TPAMI}, 35\penalty0 (11):\penalty0 2765--2781, 2013.

\bibitem[Faisal et~al.(2019)Faisal, Akhter, Ali, and Hartley]{Faisal2020wacv_eponet}
Muhammad Faisal, Ijaz Akhter, Mohsen Ali, and Richard Hartley.
\newblock Epo-net: Exploiting geometric constraints on dense trajectories for motion saliency.
\newblock In \emph{Proceedings of the IEEE/CVF Winter Conference on Applications of Computer Vision (WACV)}, pages 1873--1882, 2019.

\bibitem[Geiger et~al.(2013)Geiger, Lenz, Stiller, and Urtasun]{Geiger2013IJRR}
Andreas Geiger, Philip Lenz, Christoph Stiller, and Raquel Urtasun.
\newblock Vision meets robotics: The kitti dataset.
\newblock \emph{Int. J. Robotics Res.}, 32\penalty0 (11):\penalty0 1231--1237, 2013.

\bibitem[Greff et~al.(2022)Greff, Belletti, Beyer, Doersch, Du, Duckworth, Fleet, Gnanapragasam, Golemo, Herrmann, et~al.]{greff2021kubric}
Klaus Greff, Francois Belletti, Lucas Beyer, Carl Doersch, Yilun Du, Daniel Duckworth, David~J Fleet, Dan Gnanapragasam, Florian Golemo, Charles Herrmann, et~al.
\newblock Kubric: a scalable dataset generator.
\newblock In \emph{Proceedings of the IEEE/CVF Conference on Computer Vision and Pattern Recognition (CVPR)}, pages 3739--3751. IEEE, 2022.

\bibitem[He et~al.(2026)He, Lin, Cui, Guo, Shen, and Zhang]{he2026geomotion}
Xiankang He, Peile Lin, Ying Cui, Dongyan Guo, Chunhua Shen, and Xiaoqin Zhang.
\newblock Geomotion: Rethinking motion segmentation via latent 4d geometry.
\newblock \emph{arXiv preprint arXiv:2602.21810}, 2026.

\bibitem[Homeyer and Schn{\"o}rr(2023)]{Homeyer2023iccvw}
Christian Homeyer and Christoph Schn{\"o}rr.
\newblock On moving object segmentation from monocular video with transformers.
\newblock In \emph{2023 IEEE/CVF International Conference on Computer Vision Workshops (ICCVW)}, pages 880--891. IEEE, 2023.

\bibitem[Huang et~al.(2025{\natexlab{a}})Huang, Zhou, Rabeti, Korovko, Ling, Ren, Shen, Gao, Slepichev, Lin, et~al.]{huang2025vipe}
Jiahui Huang, Qunjie Zhou, Hesam Rabeti, Aleksandr Korovko, Huan Ling, Xuanchi Ren, Tianchang Shen, Jun Gao, Dmitry Slepichev, Chen-Hsuan Lin, et~al.
\newblock Vipe: Video pose engine for 3d geometric perception.
\newblock \emph{arXiv.org}, 2025{\natexlab{a}}.

\bibitem[Huang et~al.(2024{\natexlab{a}})Huang, Wei, Zheng, An, Lu, Zhan, Tomizuka, Keutzer, and Zhang]{huang2024sgaussian}
Nan Huang, Xiaobao Wei, Wenzhao Zheng, Pengju An, Ming Lu, Wei Zhan, Masayoshi Tomizuka, Kurt Keutzer, and Shanghang Zhang.
\newblock Sgaussian: Self-supervised street gaussians for autonomous driving.
\newblock \emph{CoRR}, 2024{\natexlab{a}}.

\bibitem[Huang et~al.(2025{\natexlab{b}})Huang, Zheng, Xu, Keutzer, Zhang, Kanazawa, and Wang]{huang2025segment}
Nan Huang, Wenzhao Zheng, Chenfeng Xu, Kurt Keutzer, Shanghang Zhang, Angjoo Kanazawa, and Qianqian Wang.
\newblock Segment any motion in videos.
\newblock In \emph{Proceedings of the IEEE/CVF Conference on Computer Vision and Pattern Recognition (CVPR)}, pages 3406--3416. IEEE, 2025{\natexlab{b}}.

\bibitem[Huang et~al.(2024{\natexlab{b}})Huang, Chen, and Zelek]{huang2024zero}
Yuxiang Huang, Yuhao Chen, and John Zelek.
\newblock Zero-shot monocular motion segmentation in the wild by combining deep learning with geometric motion model fusion.
\newblock In \emph{Proceedings of the IEEE/CVF Conference on Computer Vision and Pattern Recognition Workshops (CVPRW)}, pages 2733--2743. IEEE, 2024{\natexlab{b}}.

\bibitem[Isack and Boykov(2012)]{Isack2012ijcv}
Hossam Isack and Yuri Boykov.
\newblock Energy-based geometric multi-model fitting.
\newblock \emph{International Journal of Computer Vision}, 97\penalty0 (2):\penalty0 123--147, 2012.

\bibitem[Jadon(2020)]{jadon2020survey-dice-loss}
Shruti Jadon.
\newblock A survey of loss functions for semantic segmentation.
\newblock In \emph{IEEE Symposium on Computational Intelligence in Bioinformatics and Computational Biology}, pages 1--7. IEEE, IEEE, 2020.

\bibitem[Jocher et~al.(2023)Jocher, Chaurasia, and Qiu]{ultralytics}
Glenn Jocher, Ayush Chaurasia, and Jing Qiu.
\newblock {Ultralytics YOLO}, 2023.

\bibitem[Jun et~al.(2022)Jun, Lee, Lee, and Kim]{jun2022depth}
Jinyoung Jun, Jae-Han Lee, Chul Lee, and Chang-Su Kim.
\newblock Depth map decomposition for monocular depth estimation.
\newblock In \emph{Proceedings of the European Conference on Computer Vision (ECCV)}, pages 18--34. Springer, Springer Nature Switzerland, 2022.

\bibitem[Karaev et~al.(2023)Karaev, Rocco, Graham, Neverova, Vedaldi, and Rupprecht]{karaev2023dynamicstereo}
Nikita Karaev, Ignacio Rocco, Benjamin Graham, Natalia Neverova, Andrea Vedaldi, and Christian Rupprecht.
\newblock Dynamicstereo: Consistent dynamic depth from stereo videos.
\newblock In \emph{Proceedings of the IEEE/CVF Conference on Computer Vision and Pattern Recognition (CVPR)}, pages 13229--13239. IEEE, 2023.

\bibitem[Karazija et~al.(2024)Karazija, Laina, Rupprecht, and Vedaldi]{karazija24learning}
Laurynas Karazija, Iro Laina, Christian Rupprecht, and Andrea Vedaldi.
\newblock Learning segmentation from point trajectories.
\newblock \emph{Advances in Neural Information Processing Systems}, 37:\penalty0 112573--112597, 2024.

\bibitem[Kong and Fu(2022)]{kong2022human}
Yu Kong and Yun Fu.
\newblock Human action recognition and prediction: A survey.
\newblock \emph{IJCV}, 130\penalty0 (5):\penalty0 1366--1401, 2022.

\bibitem[Lee et~al.(2019)Lee, Im, Lin, and Kweon]{lee2019learning}
Seokju Lee, Sunghoon Im, Stephen Lin, and In~So Kweon.
\newblock Learning residual flow as dynamic motion from stereo videos.
\newblock In \emph{2019 IEEE/RSJ International Conference on Intelligent Robots and Systems (IROS)}, pages 1180--1186, 2019.

\bibitem[Li et~al.(2013)Li, Kim, Humayun, Tsai, and Rehg]{li2013video}
Fuxin Li, Taeyoung Kim, Ahmad Humayun, David Tsai, and James~M Rehg.
\newblock Video segmentation by tracking many figure-ground segments.
\newblock In \emph{Proceedings of the IEEE/CVF International Conference on Computer Vision (ICCV)}, pages 2192--2199. IEEE, 2013.

\bibitem[Meunier et~al.(2023)Meunier, Badoual, and Bouthemy]{EM}
Etienne Meunier, Anaïs Badoual, and Patrick Bouthemy.
\newblock Em-driven unsupervised learning for efficient motion segmentation.
\newblock \emph{IEEE TPAMI}, 45\penalty0 (4):\penalty0 4462--4473, 2023.

\bibitem[Mohamed et~al.(2020)Mohamed, Ewaisha, Siam, Rashed, Yogamani, Hamdy, El-Dakdouky, and Sallab]{mohamed2021monocular}
Eslam Mohamed, Mahmoud Ewaisha, M. Siam, Hazem Rashed, S. Yogamani, Waleed Hamdy, Mohamed El-Dakdouky, and Ahmad~El Sallab.
\newblock Monocular instance motion segmentation for autonomous driving: Kitti instancemotseg dataset and multi-task baseline.
\newblock In \emph{2021 IEEE Intelligent Vehicles Symposium (IV)}, pages 114--121. IEEE, 2020.

\bibitem[Morris et~al.(2025)Morris, Wang, Kliniewski, and Ila]{morris2025dynosam}
Jesse Morris, Yiduo Wang, Mikolaj Kliniewski, and Viorela Ila.
\newblock Dynosam: Open-source smoothing and mapping framework for dynamic slam.
\newblock \emph{IEEE Transactions on robotics}, 2025.

\bibitem[Neoral et~al.(2021)Neoral, Sochman, and Matas]{Neoral2021bmvc}
Michal Neoral, Jan Sochman, and Jiri Matas.
\newblock Monocular arbitrary moving object discovery and segmentation.
\newblock In \emph{BMVC}. British Machine Vision Association, 2021.

\bibitem[Ochs et~al.(2014)Ochs, Malik, and Brox]{Ochs2014pami}
Peter Ochs, Jitendra Malik, and Thomas Brox.
\newblock Segmentation of moving objects by long term video analysis.
\newblock \emph{IEEE TPAMI}, 36\penalty0 (6):\penalty0 1187--1200, 2014.

\bibitem[Oquab et~al.(2023)Oquab, Darcet, Moutakanni, Vo, Szafraniec, Khalidov, Fernandez, Haziza, Massa, El-Nouby, et~al.]{oquab2023dinov2}
Maxime Oquab, Timothée Darcet, Theo Moutakanni, Huy~V. Vo, Marc Szafraniec, Vasil Khalidov, Pierre Fernandez, Daniel Haziza, Francisco Massa, Alaaeldin El-Nouby, et~al.
\newblock Dinov2: Learning robust visual features without supervision.
\newblock \emph{Trans. Mach. Learn. Res.}, 2023.

\bibitem[Papazoglou and Ferrari(2013)]{Papazoglou2013iccv}
Anestis Papazoglou and Vittorio Ferrari.
\newblock Fast object segmentation in unconstrained video.
\newblock In \emph{ICCV}, pages 1777--1784, 2013.

\bibitem[Perazzi et~al.(2016)Perazzi, Pont-Tuset, McWilliams, Gool, Gross, and Sorkine-Hornung]{perazzi2016benchmark}
Federico Perazzi, J. Pont-Tuset, B. McWilliams, L. Gool, M. Gross, and A. Sorkine-Hornung.
\newblock A benchmark dataset and evaluation methodology for video object segmentation.
\newblock In \emph{CVPR}, pages 724--732. IEEE, 2016.

\bibitem[Rao et~al.(2010)Rao, Tron, Vidal, and Ma]{Rao2010pami}
Shankar Rao, Roberto Tron, Rene Vidal, and Yi Ma.
\newblock Motion segmentation in the presence of outlying, incomplete, or corrupted trajectories.
\newblock \emph{IEEE TPAMI}, 32\penalty0 (10):\penalty0 1832--1845, 2010.

\bibitem[Rashed et~al.(2019)Rashed, El~Sallab, Yogamani, and ElHelw]{rashed2019motion}
Hazem Rashed, Ahmad El~Sallab, Senthil Yogamani, and Mohamed ElHelw.
\newblock Motion and depth augmented semantic segmentation for autonomous navigation.
\newblock In \emph{2019 IEEE/CVF Conference on Computer Vision and Pattern Recognition Workshops (CVPRW)}, pages 364--370. IEEE, 2019.

\bibitem[Ravi et~al.(2024)Ravi, Gabeur, Hu, Hu, Ryali, Ma, Khedr, Rädle, Rolland, Gustafson, et~al.]{ravi2024sam2}
Nikhila Ravi, Valentin Gabeur, Yuan-Ting Hu, Ronghang Hu, Chaitanya~K. Ryali, Tengyu Ma, Haitham Khedr, Roman Rädle, Chloé Rolland, Laura Gustafson, et~al.
\newblock Sam 2: Segment anything in images and videos.
\newblock \emph{ICCV}, 2024.

\bibitem[Sekkati and Mitiche(2007)]{Sekkati2007ras}
Hicham Sekkati and Amar Mitiche.
\newblock A variational method for the recovery of dense 3d structure from motion.
\newblock \emph{Robotics Auton. Syst.}, 55\penalty0 (7):\penalty0 597--607, 2007.

\bibitem[Taniai et~al.(2017)Taniai, Sinha, and Sato]{taniai2017fast}
Tatsunori Taniai, Sudipta~N. Sinha, and Yoichi Sato.
\newblock Fast multi-frame stereo scene flow with motion segmentation.
\newblock In \emph{Proceedings of the IEEE Conference on Computer Vision and Pattern Recognition (CVPR)}, 2017.

\bibitem[Teed and Deng(2020)]{teed2020raft}
Zachary Teed and Jia Deng.
\newblock Raft: Recurrent all-pairs field transforms for optical flow.
\newblock In \emph{ECCV}, pages 402--419. Springer International Publishing, 2020.

\bibitem[Tokmakov et~al.(2019)Tokmakov, Schmid, and Alahari]{Tokmakov2019ijcv}
Pavel Tokmakov, Cordelia Schmid, and Karteek Alahari.
\newblock Learning to segment moving objects.
\newblock \emph{International Journal of Computer Vision}, 127\penalty0 (3):\penalty0 282–301, 2019.

\bibitem[Wang et~al.(2025)Wang, Morris, Wu, Vidal-Calleja, and Ila]{Wang2025icra_dynorecon}
Yiduo Wang, Jesse Morris, Lan Wu, Teresa Vidal-Calleja, and Viorela Ila.
\newblock Dynorecon: Dynamic object reconstruction for navigation.
\newblock In \emph{2025 IEEE International Conference on Robotics and Automation (ICRA)}, pages 3305--3311. IEEE, 2025.

\bibitem[Wang et~al.(2026)Wang, Turner, and Ila]{wang2026linstereo}
Yiran Wang, Oliver Turner, and Viorela Ila.
\newblock Linstereo: linear-complexity global attention for multi-scale iterative stereo matching.
\newblock In \emph{European Conference on Computer Vision}, pages 1--20. Springer, 2026.

\bibitem[Wedel et~al.(2009)Wedel, Meißner, Rabe, Franke, and Cremers]{wedel2009detection}
A. Wedel, Annemarie Meißner, C. Rabe, Uwe Franke, and D. Cremers.
\newblock Detection and segmentation of independently moving objects from dense scene flow.
\newblock In \emph{Energy Minimization Methods in Computer Vision and Pattern Recognition}, pages 14--27. Springer, Springer Berlin Heidelberg, 2009.

\bibitem[Wei et~al.(2024)Wei, Chen, Lu, Chen, and Tian]{wei2025graphavatar}
Xiaobao Wei, Peng Chen, Ming Lu, Hui Chen, and Feng Tian.
\newblock Graphavatar: Compact head avatars with gnn-generated 3d gaussians.
\newblock In \emph{AAAI Conference on Artificial Intelligence}, pages 8295--8303, 2024.

\bibitem[Wen et~al.(2025)Wen, Trepte, Aribido, Kautz, Gallo, and Birchfield]{wen2025stereo}
Bowen Wen, Matthew Trepte, Joseph Aribido, Jan Kautz, Orazio Gallo, and Stan Birchfield.
\newblock Foundationstereo: Zero-shot stereo matching.
\newblock \emph{CVPR}, 2025.

\bibitem[Woo et~al.(2018)Woo, Park, Lee, and Kweon]{woo2018cbam}
Sanghyun Woo, Jongchan Park, Joon-Young Lee, and In~So Kweon.
\newblock Cbam: Convolutional block attention module.
\newblock In \emph{ECCV}, pages 3--19. Springer International Publishing, 2018.

\bibitem[Xie et~al.(2022)Xie, Xie, and Zisserman]{OCLR}
Junyu Xie, Weidi Xie, and Andrew Zisserman.
\newblock Segmenting moving objects via an object-centric layered representation.
\newblock In \emph{NeurIPS}, pages 28023--28036. Neural Information Processing Systems Foundation, Inc. (NeurIPS), 2022.

\bibitem[Xie et~al.(2023)Xie, Xie, and Zisserman]{xie24appearrefine}
Junyu Xie, Weidi Xie, and Andrew Zisserman.
\newblock Appearance-based refinement for object-centric motion segmentation.
\newblock In \emph{ECCV}, 2023.

\bibitem[Xie et~al.(2024)Xie, Yang, Xie, and Zisserman]{xie2024flowsam}
Junyu Xie, Charig Yang, Weidi Xie, and Andrew Zisserman.
\newblock Moving object segmentation: All you need is sam (and flow).
\newblock In \emph{ACCV}, 2024.

\bibitem[Xie et~al.(2026)Xie, Han, Xie, and Zisserman]{xie2026gmos}
Junyu Xie, Tengda Han, Weidi Xie, and Andrew Zisserman.
\newblock Gmos: Grounding moving object segmentation in 3d space and time.
\newblock \emph{arXiv preprint arXiv:2605.30352}, 2026.

\bibitem[Xu et~al.(2018{\natexlab{a}})Xu, Yang, Fan, Yue, Liang, Yang, and Huang]{vos2018}
Ning Xu, Linjie Yang, Yuchen Fan, Dingcheng Yue, Yuchen Liang, Jianchao Yang, and Thomas~S. Huang.
\newblock Youtube-vos: {A} large-scale video object segmentation benchmark.
\newblock \emph{arXiv.org}, abs/1809.03327, 2018{\natexlab{a}}.

\bibitem[Xu et~al.(2018{\natexlab{b}})Xu, Cheong, and Li]{xu2018motion}
Xun Xu, Loong~Fah Cheong, and Zhuwen Li.
\newblock Motion segmentation by exploiting complementary geometric models.
\newblock In \emph{CVPR}, pages 2859--2867. IEEE, 2018{\natexlab{b}}.

\bibitem[Yang et~al.(2024)Yang, Kang, Huang, Zhao, Xu, Feng, and Zhao]{depth_anything_v2}
Lihe Yang, Bingyi Kang, Zilong Huang, Zhen Zhao, Xiaogang Xu, Jiashi Feng, and Hengshuang Zhao.
\newblock Depth anything v2.
\newblock \emph{Advances in Neural Information Processing Systems}, 37:\penalty0 21875--21911, 2024.

\bibitem[Yao et~al.(2020)Yao, Lin, Xia, Zhao, and Zhou]{video_object_segmentation_survey_yao2020video}
Rui Yao, Guosheng Lin, Shixiong Xia, Jiaqi Zhao, and Yong Zhou.
\newblock Video object segmentation and tracking: A survey.
\newblock \emph{ACM Transactions on Intelligent Systems and Technology (TIST)}, 11\penalty0 (4):\penalty0 1--47, 2020.

\bibitem[Yu et~al.(2016)Yu, Jiang, Wang, Cao, and Huang]{yu2016unitbox-iouloss}
Jiahui Yu, Yuning Jiang, Zhangyang Wang, Zhimin Cao, and Thomas Huang.
\newblock Unitbox: An advanced object detection network.
\newblock In \emph{ACM Multimedia}, pages 516--520. IEEE, 2016.

\bibitem[Zhong et~al.(2025)Zhong, Jin, Popovi{\'c}, Behley, and Stachniss]{pi3_zhong2025dynamic}
Xingguang Zhong, Liren Jin, Marija Popovi{\'c}, Jens Behley, and Cyrill Stachniss.
\newblock Dynamic visual slam using a general 3d prior.
\newblock \emph{arXiv preprint}, 2025.

\bibitem[Zhou et~al.(2020)Zhou, Wang, Zhou, Yao, Li, and Shao]{Zhou2020aaai}
Tianfei Zhou, Shunzhou Wang, Yi Zhou, Yazhou Yao, Jianwu Li, and Ling Shao.
\newblock Motion-attentive transition for zero-shot video object segmentation.
\newblock In \emph{AAAI}, pages 13066--13073, 2020.

\bibitem[Zhou et~al.(2024)Zhou, Wu, Boutteau, Yang, and Ginhac]{zhou2024event}
Zhuyun Zhou, Zongwei Wu, Rémi Boutteau, Fan Yang, and D. Ginhac.
\newblock Event-free moving object segmentation from moving ego vehicle.
\newblock pages 8960--8965. IEEE, 2024.

\bibitem[Zou et~al.(2018)Zou, Luo, and Huang]{zou2018dfnet}
Yuliang Zou, Zelun Luo, and Jia-Bin Huang.
\newblock Df-net: Unsupervised joint learning of depth and flow using cross-task consistency.
\newblock In \emph{Proceedings of the European Conference on Computer Vision (ECCV)}, 2018.

\end{thebibliography}
}

\clearpage
\appendix
\numberwithin{figure}{section}
\numberwithin{table}{section}
\numberwithin{equation}{section}
\twocolumn[
\begin{center}
    {\LARGE\bfseries Supplementary Materials\par}
    \vspace{0.8em}
    {\Large\bfseries DynOSeg: Dynamic Object Segmentation via Motion-Geometry Fusion\par}
    \vspace{24pt}
    {\large\lineskip .5em\begin{tabular}[t]{c}\paperauthors\end{tabular}\par}
    \vspace{1.5em}
\end{center}
]
\section{Qualitative Results}
\label{appendix:qualitative_results}
\begin{figure*}[t]
    \centering
    \includegraphics[width=\linewidth]{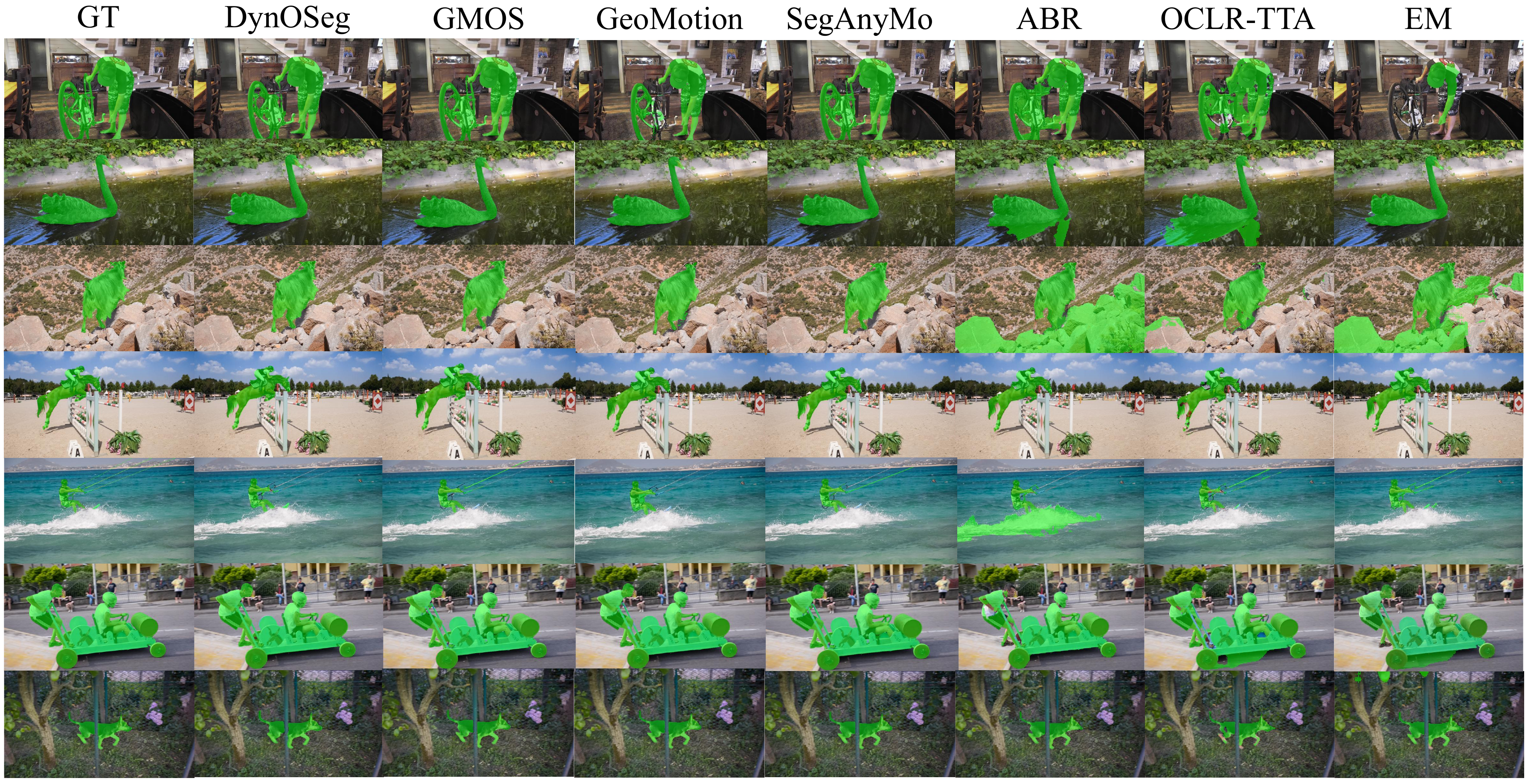}
    \caption{\textbf{Qualitative comparison on DAVIS2017-M.} Columns show the ground truth and predictions from DynOSeg, GMOS, GeoMotion, SegAnyMo, ABR, OCLR-TTA, and EM. Green overlays indicate moving-object foreground regions.}
    \label{fig:supp_davis_qualitative}
\end{figure*}

\begin{table}[t]
\centering
\caption{\textbf{PlainCNN encoder architecture.} $C_{\mathrm{in}}=4$ for depth plus normals and $C_{\mathrm{in}}=1$ for the normalized residual map. Only the numbered convolutional blocks count toward the four-layer depth.}
\label{tab:plaincnn_architecture}
\footnotesize
\setlength{\tabcolsep}{3pt}
\renewcommand{\arraystretch}{1.08}
\begin{tabular*}{\columnwidth}{@{\extracolsep{\fill}}clcc@{}}
\toprule
Conv. & Operation & Channels (in/out) & Kernel / stride \\
\midrule
1 & Conv--BN--SiLU & $C_{\mathrm{in}}/64$ & $3\times3$ / $2$ \\
2 & Conv--BN--SiLU & $64/128$ & $3\times3$ / $1$ \\
-- & MaxPool & $128/128$ & $2\times2$ / $2$ \\
3 & Conv--BN--SiLU & $128/256$ & $3\times3$ / $1$ \\
4 & Conv--BN--SiLU & $256/256$ & $3\times3$ / $1$ \\
-- & LayerNorm2d & $256/256$ & -- \\
-- & Dropout ($p=0.1$) & $256/256$ & -- \\
\bottomrule
\end{tabular*}
\end{table}

\paragraph{DAVIS2017-M.}
Figure~\ref{fig:supp_davis_qualitative} compares foreground masks produced by DynOSeg and the evaluated baselines on representative DAVIS2017-M frames.
Green overlays denote the ground-truth or predicted moving-object regions.
In these examples, DynOSeg preserves object extent and boundary detail across articulated motion, thin structures, scale changes, and cluttered backgrounds, whereas competing methods sometimes omit object parts or include nearby background regions.

\begin{figure*}[t]
    \centering
    \includegraphics[width=\linewidth]{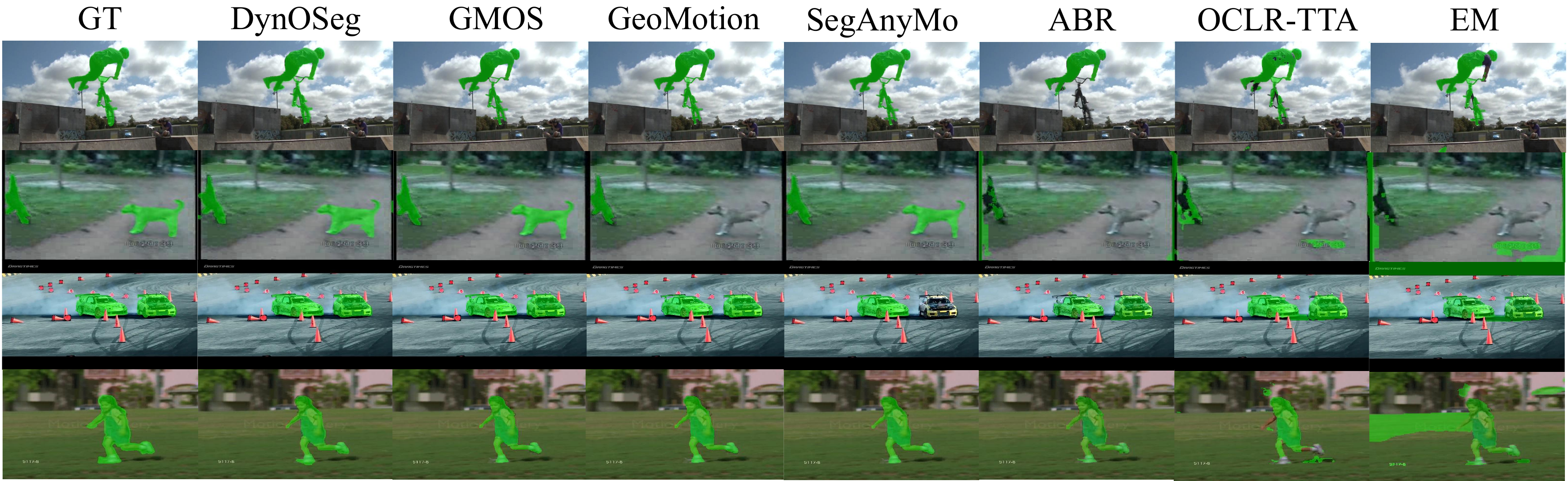}
    \caption{\textbf{Qualitative comparison on SegTrackV2.} Columns show the ground truth and predictions from DynOSeg, GMOS, GeoMotion, SegAnyMo, ABR, OCLR-TTA, and EM. Green overlays indicate moving-object foreground regions.}
    \label{fig:supp_segtrack_qualitative}
\end{figure*}

\paragraph{SegTrackV2.}
Figure~\ref{fig:supp_segtrack_qualitative} shows examples containing multiple targets, articulated motion, motion blur, and large appearance changes.
DynOSeg retains separate moving regions and follows the visible object silhouettes in these selected frames.
Several baselines either miss a target or part of its body, while others introduce foreground responses in nearby static regions.

\begin{figure*}[t]
    \centering
    \includegraphics[width=\linewidth]{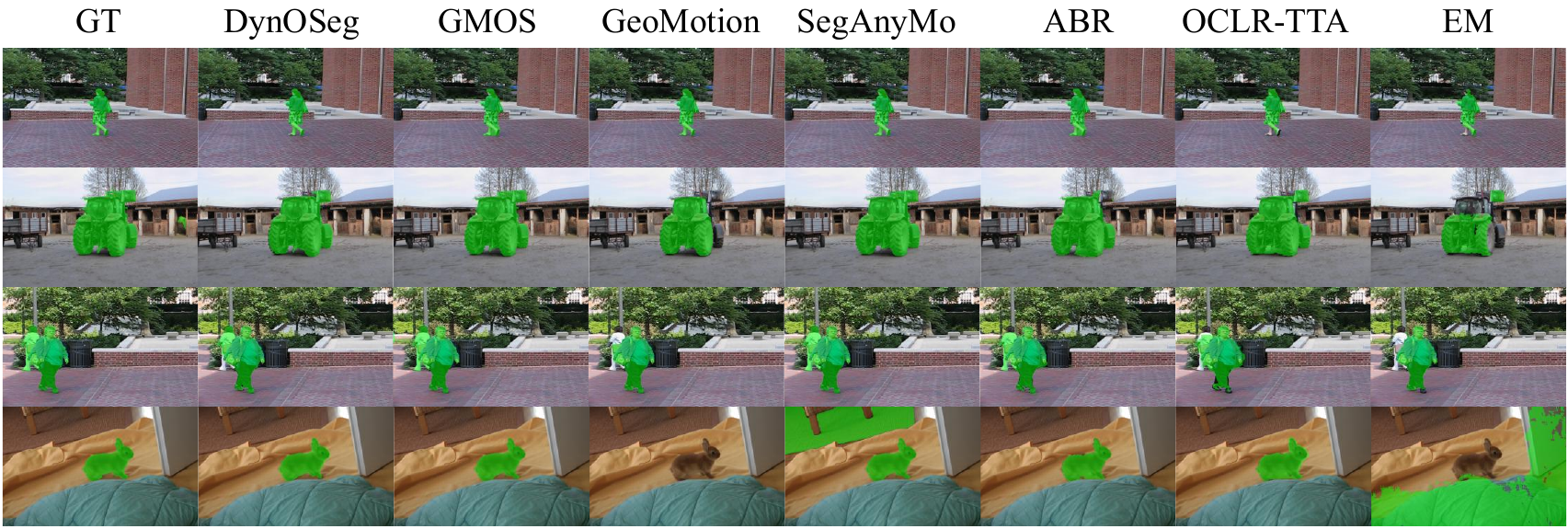}
    \caption{\textbf{Qualitative comparison on FBMS-59.} Columns show the ground truth and predictions from DynOSeg, GMOS, GeoMotion, SegAnyMo, ABR, OCLR-TTA, and EM. Green overlays indicate moving-object foreground regions.}
    \label{fig:supp_fbms_qualitative}
\end{figure*}

\paragraph{FBMS-59.}
Figure~\ref{fig:supp_fbms_qualitative} includes moving people, a large vehicle, and a small animal under substantial scale and appearance variation.
DynOSeg produces compact masks that remain aligned with the moving targets in these examples.
By comparison, some methods miss small targets or absorb adjacent static background into the prediction.

\begin{figure*}[t]
    \centering
    \includegraphics[width=\linewidth]{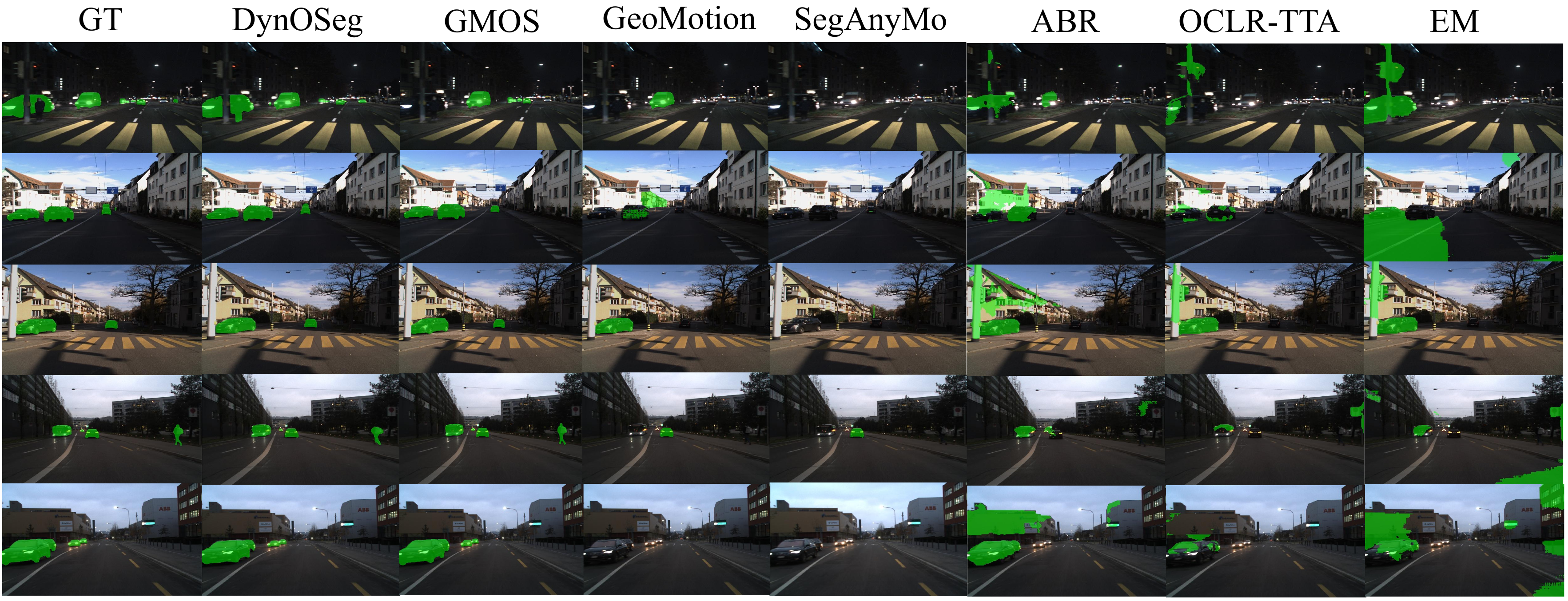}
    \caption{\textbf{Qualitative comparison on DSEC-MOS.} Columns show the ground truth and predictions from DynOSeg, GMOS, GeoMotion, SegAnyMo, ABR, OCLR-TTA, and EM. Green overlays indicate moving-object foreground regions.}
    \label{fig:supp_dsec_qualitative}
\end{figure*}

\paragraph{DSEC-MOS.}
Figure~\ref{fig:supp_dsec_qualitative} presents urban-driving scenes with strong ego-motion, low illumination, and small or distant road users.
DynOSeg identifies multiple moving vehicles and pedestrians while limiting responses on the road and surrounding structures in the displayed frames.
Several baselines omit distant targets or produce large false-positive regions in the static background.
Overall, these selected visualizations complement the quantitative results in Table~2 of the main paper; they are not intended to establish a dataset-wide ranking on their own.

\subsection{Ablation Visualizations}
\label{appendix:Ablation}
\begin{figure*}[t]
    \centering
    \includegraphics[width=0.82\linewidth]{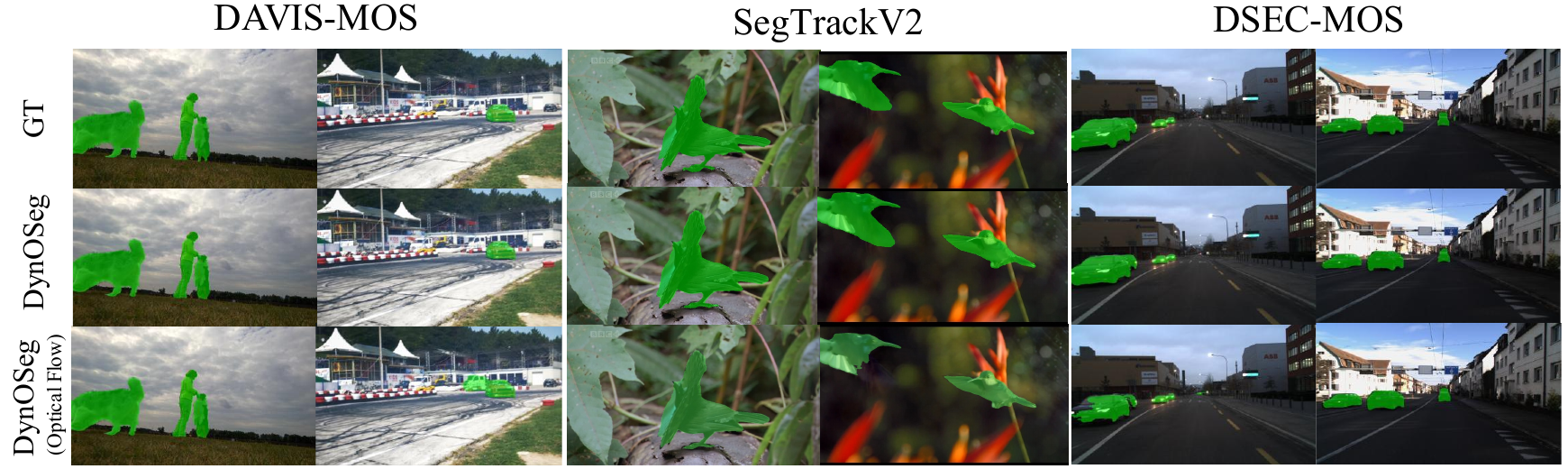}
    \caption{\textbf{Residual-map versus optical-flow input.} Rows show the ground truth, DynOSeg with the residual map, and an otherwise matched DynOSeg variant using raw optical flow. Columns contain examples from DAVIS2017-M and DSEC-MOS. Green overlays denote moving-object foreground regions.}
    \label{fig:supp_residual_vs_flow}
\end{figure*}

\paragraph{Residual map versus optical flow.}
Figure~\ref{fig:supp_residual_vs_flow} visualizes the comparison between rows (A) and (F) of Table~4 in the main paper.
Both variants use RGB and temporally consistent ViPE depth; only the motion representation changes.
In these examples, the residual-map variant better preserves complete foreground regions and small or distant moving objects, whereas raw optical flow is more affected by apparent motion induced by the camera.

\begin{figure*}[t]
    \centering
    \includegraphics[width=\linewidth]{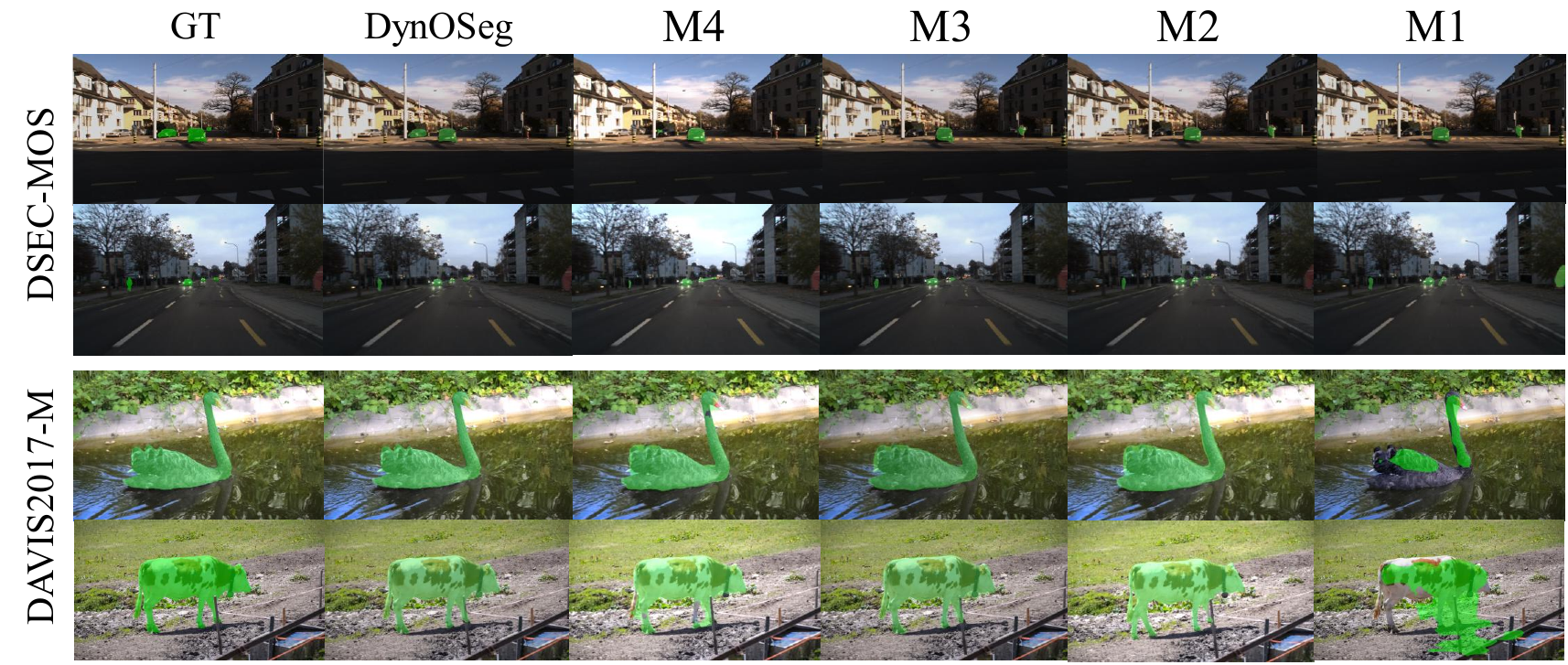}
    \caption{\textbf{Qualitative ablation of input modalities.} M1 uses only the residual map; M2 uses the residual map and ViPE depth; M3 uses the residual map and RGB; and M4 uses the residual map, RGB, and per-frame Depth Anything V2 depth. DynOSeg uses the residual map, RGB, and temporally consistent ViPE depth. Rows show DSEC-MOS and DAVIS2017-M examples, and green overlays denote predicted or ground-truth foreground.}
    \label{fig:supp_different_inputs}
\end{figure*}

\paragraph{Different input combinations.}
Figure~\ref{fig:supp_different_inputs} complements the input-modality ablation in Table~4 of the main paper.
Using the residual map alone (M1) can miss substantial object regions, while adding ViPE depth (M2) or RGB (M3) improves object extent and boundary coherence in the displayed frames.
M4 shows that per-frame Depth Anything V2 depth does not provide the same benefit as temporally consistent ViPE depth.
The full DynOSeg configuration combines the residual map, RGB, and ViPE depth and most closely follows the ground-truth masks in these selected examples.

\begin{figure*}[t]
    \centering
    \includegraphics[width=\linewidth]{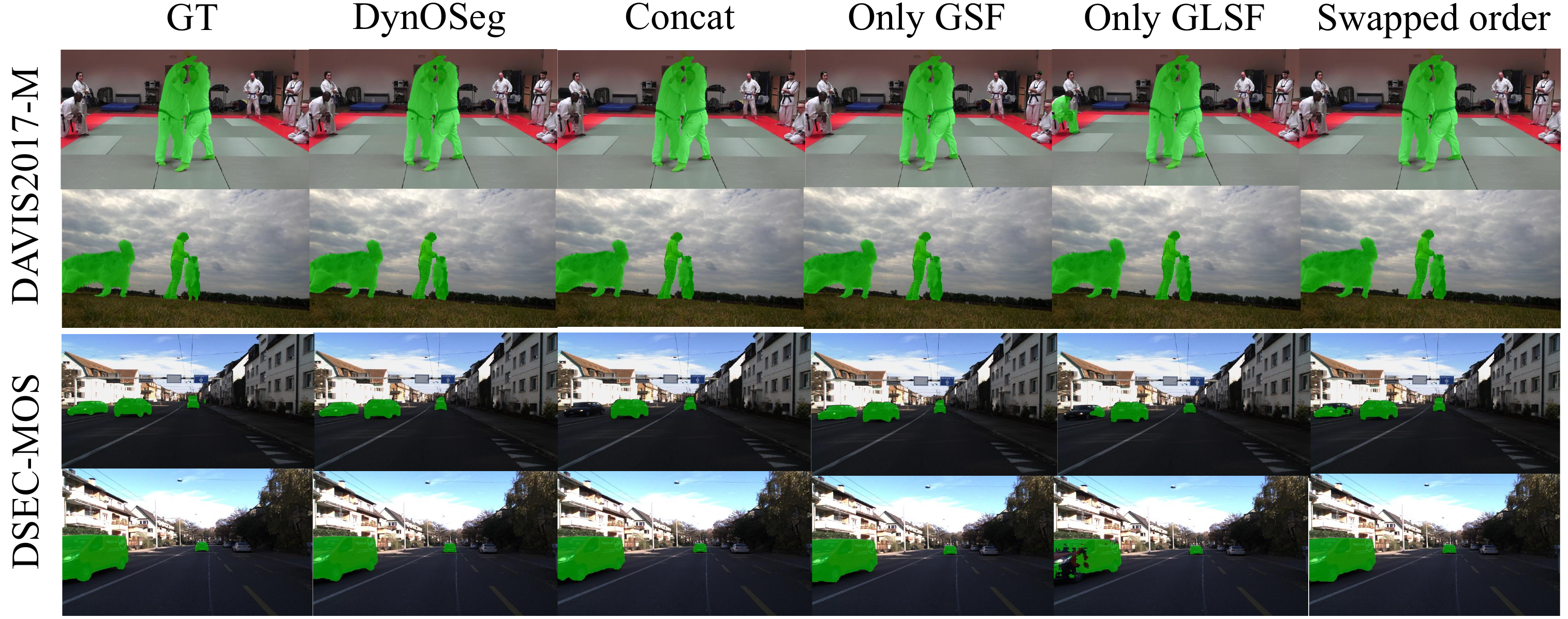}
    \caption{\textbf{Qualitative ablation of fusion structures.} Columns compare ground truth, the full two-stage DynOSeg model, direct feature concatenation, GSF only, GLSF only, and the swapped GLSF--GSF order. Rows show examples from DAVIS2017-M and DSEC-MOS. Green overlays denote moving-object foreground regions.}
    \label{fig:supp_fusion_ablation}
\end{figure*}

\paragraph{Fusion structures.}
Figure~\ref{fig:supp_fusion_ablation} visualizes the fusion ablation reported in Table~6 of the main paper with identical residual, depth, and RGB inputs.
The alternative designs can omit object parts, introduce nearby background, or fragment small foreground regions in the displayed examples.
The full two-stage GSF--GLSF design produces more coherent masks, illustrating how geometry--motion fusion followed by semantic refinement complements the quantitative results.

The ablation figures show representative cases and are intended to explain characteristic differences between variants rather than establish a dataset-wide ranking independently of the quantitative tables.

\section{CNN Baseline for Geometry Encoders}
\label{appendix:CNN-Encoders}
\label{sec:cnn_geometry_baseline}

\paragraph{Replacement protocol.}
Table~5 in the main paper separates four configurations: replacing only the residual encoder, replacing only the depth encoder, replacing both, and retaining unmodified DynOSeg.
Each selected encoder is replaced with a separate PlainCNN branch; an unselected encoder retains its original design.
The replacement does not add a CNN to the original encoders or replace the frozen RGB DINOv2 encoder.
The depth branch retains the four-channel depth-plus-normals input, and the residual branch retains its normalized single-channel residual-map input.
Both branches output $256$ channels at an output stride of $4$.
Replacing the depth encoder removes its C2f blocks; replacing the residual encoder removes its C2f blocks and spatial attention.
GSF, GLSF, DINOv2, the decoder, loss, and training settings are unchanged.

\paragraph{Architecture and layer counting.}
Table~\ref{tab:plaincnn_architecture} gives the complete layer sequence.
Each convolution uses the same Conv--BN--SiLU implementation as the proposed model: convolution followed by batch normalization and the SiLU activation.
The $2\times2$ max-pooling operation follows the second convolution, and the fourth convolution is followed by LayerNorm2d with $256$ channels and Dropout with probability $0.1$.
The term \emph{four-layer CNN} counts only the four convolutional layers; MaxPool, LayerNorm, Dropout, batch normalization, and activation operators are not counted as additional convolutional layers.
This is a layer-count convention, not a claim that normalization layers have no learnable parameters.

\paragraph{Parameter matching and results.}
Table~\ref{tab:cnn_encoder_params} compares the encoder parameter counts.
PlainCNN has $14{,}592$ additional depth-encoder parameters ($1.54\%$) and $14{,}494$ additional residual-encoder parameters ($1.53\%$), making the comparison closely, but not exactly, parameter-matched.
These counts refer to the depth and residual encoders, not the complete DynOSeg system.
Table~5 in the main paper reports $90.1/73.0$ $\mathcal{J}\&\mathcal{F}$ for unmodified DynOSeg on DAVIS2017-M/DSEC-MOS.
Defining $\Delta$ as the replacement score minus this reference, replacing only the residual encoder yields $86.0/65.4$ ($\Delta=-4.1/-7.6$ points), while replacing only the depth encoder yields $84.8/64.9$ ($\Delta=-5.3/-8.1$ points).
Replacing both yields $85.4/65.1$ ($\Delta=-4.7/-7.9$ points).
Despite their slightly larger parameter budgets, all three PlainCNN variants score below DynOSeg, supporting the original design of each branch in these tested configurations.
The joint replacement is $0.6/0.2$ points above the depth-only replacement, so the decreases are not additive or monotonic in the number of replaced encoders.
These point estimates alone do not establish a statistically reliable interaction between the branches.

\begin{table}[t]
\centering
\caption{\textbf{Geometry-encoder parameter comparison.} Exact counts are reported; the relative difference uses the corresponding DynOSeg encoder as the denominator.}
\label{tab:cnn_encoder_params}
\footnotesize
\setlength{\tabcolsep}{4pt}
\begin{tabular*}{\columnwidth}{@{\extracolsep{\fill}}lrrr@{}}
\toprule
Encoder branch & DynOSeg & PlainCNN & Difference \\
\midrule
Depth & 948,096 & 962,688 & $+1.54\%$ \\
Residual & 946,466 & 960,960 & $+1.53\%$ \\
\bottomrule
\end{tabular*}
\end{table}

\section{Custom-Curated Dataset: CARLA-MOS}
\label{appendix:CARLA-MOS}
\label{sec:dataset}
\begin{figure*}[t]
    \centering
    \includegraphics[width=\linewidth]{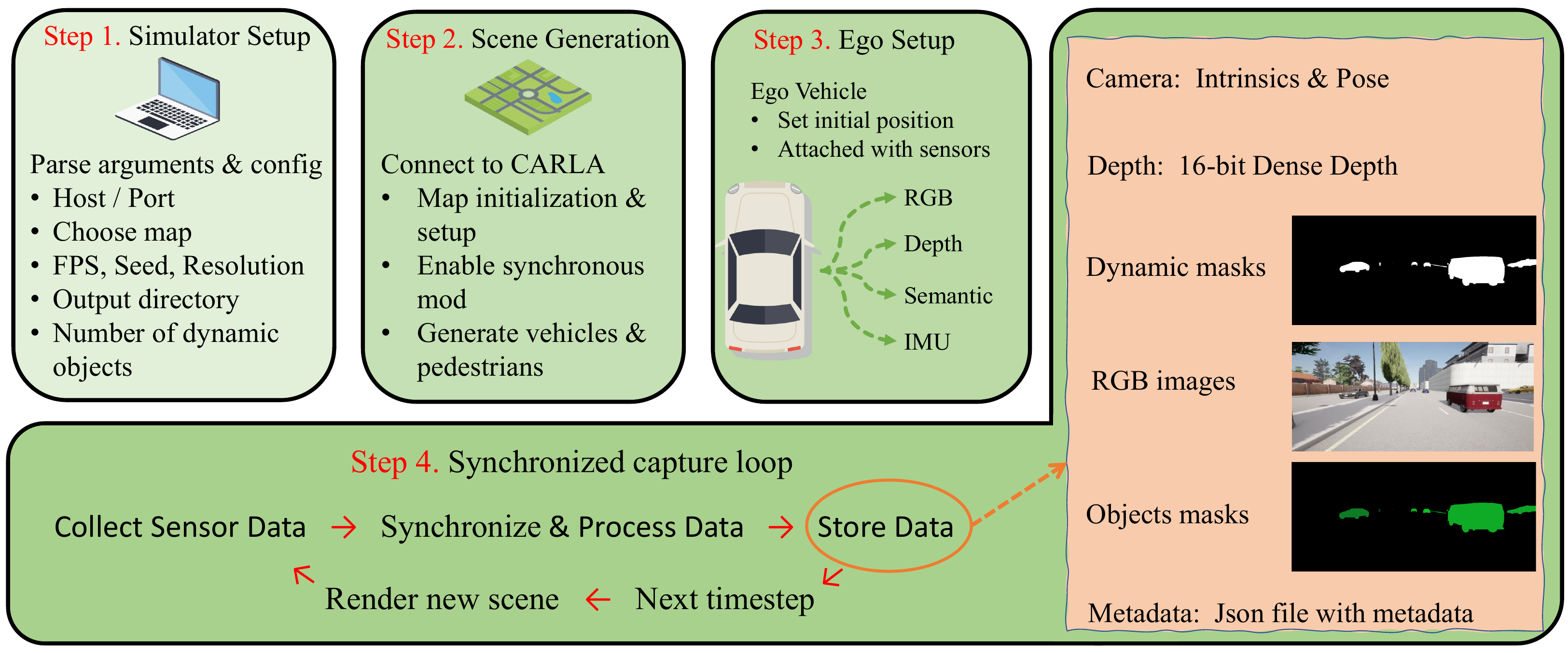}
    \caption{\textbf{CARLA-based data collection pipeline.} Simulator setup, scene generation, and ego-sensor configuration are followed by a synchronized capture loop that records RGB, semantic segmentation, depth, and camera parameters at each step; dynamic-object masks are then derived from actor motion and stored with the other modalities.}
    \label{fig:CARLA-pipline}
\end{figure*}

\begin{figure*}[t]
  \centering
  \includegraphics[width=\linewidth]{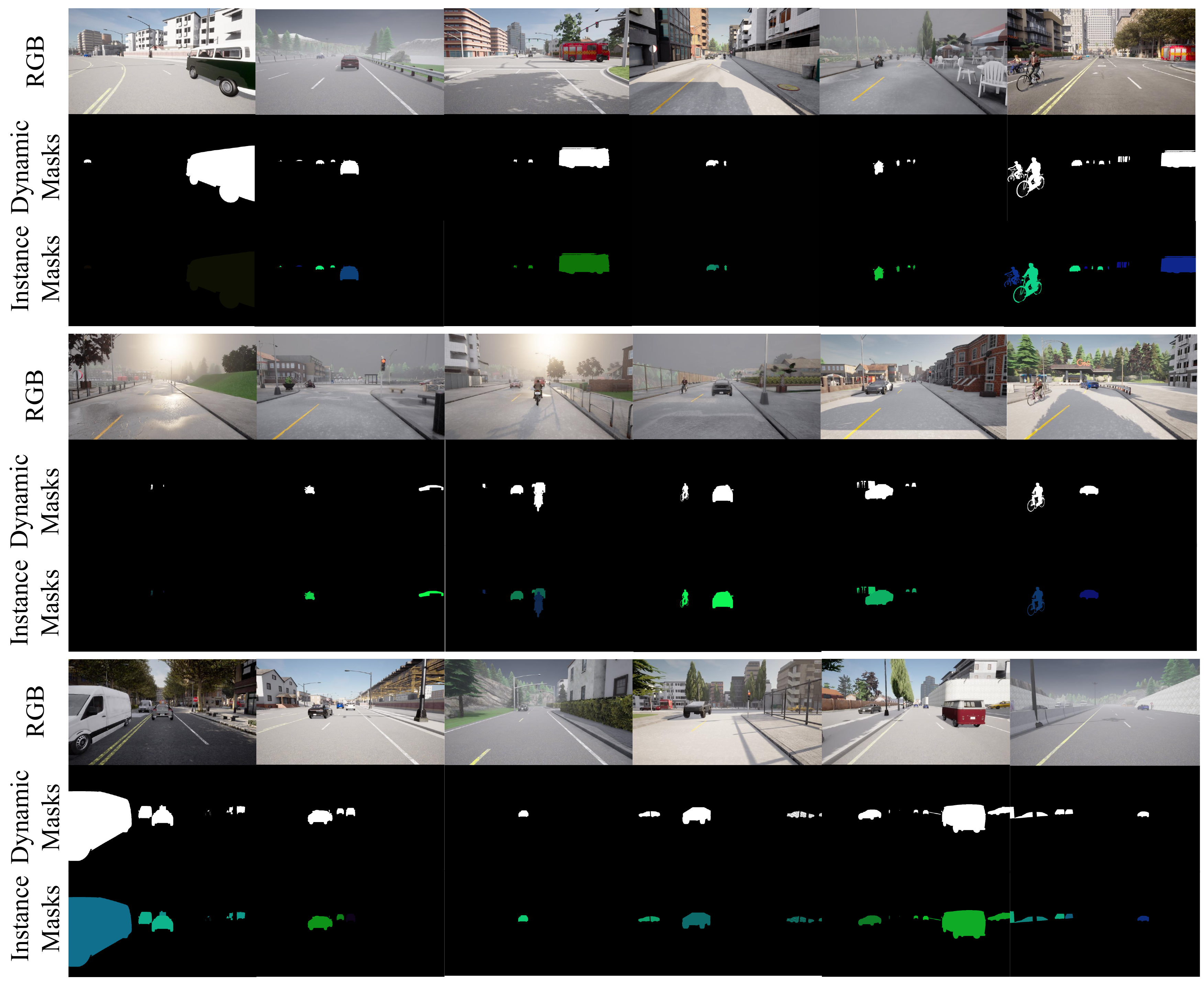}
  \caption{\textbf{CARLA-MOS:} a custom-curated simulated driving-scene MOS dataset generated with CARLA~\cite{Dosovitskiy17} for DynOSeg training.}
  \label{fig:sec4-carla-mos}
\end{figure*}

To further extend DynOSeg to autonomous driving scenarios, we construct a custom simulated driving dataset, termed \textit{CARLA-MOS}, using the CARLA simulator~\cite{Dosovitskiy17}. \Cref{fig:sec4-carla-mos} presents representative samples from the proposed dataset, while \cref{fig:CARLA-pipline} illustrates the overall CARLA-based dataset collection pipeline.
The dataset contains $1{,}000$ sequences and $300$k frames, all used as training data; it has no held-out test split in this study.
Its contribution is evaluated by transfer to the real-world benchmarks in Table~3 of the main paper, not by in-domain CARLA-MOS accuracy.

The pipeline follows four stages: simulator setup, scene generation, ego-sensor configuration, and a synchronized capture loop. We first configure the simulator environment and traffic manager and then spawn ego vehicles together with non-player trucks, cars, bicycles, motorcycles, and pedestrians with randomized trajectories and speeds. At each simulation step, the pipeline synchronously records RGB images, instance masks, semantic labels, depth maps, camera poses, and intrinsic parameters under varied urban traffic conditions.

Dynamic objects are identified from their motion states in the simulator, and pixel-level moving-object masks are generated from CARLA's instance annotations. The synchronized multimodal annotations also include depth observations and camera calibration parameters. We will release CARLA-MOS and the corresponding data-collection scripts with the DynOSeg source code.

The collected data are organized in a structured format for efficient training and evaluation. RGB observations are stored as JPEG images, while binary moving-object masks are saved as PNG files. Instance-level semantic segmentation annotations are also collected, where each object instance is represented using a unique RGB color encoding and a unique ID number in the metadata. Depth maps are compressed into a ZIP archive, with each frame stored as a 16-bit PNG image representing metric depth values. Camera trajectories are saved as $4 \times 4$ camera-to-world transformation matrices, and camera intrinsics are stored as a $3 \times 3$ intrinsic matrix. In addition, all simulation metadata, including timestamps, actor identities, object statistics, and simulation settings, are serialized into a metadata JSON file for reproducibility and future analysis.

\section{KITTI-MOS}
\label{appendix:KITTI-MOS}
\label{sec:kitti-mos}

\begin{table*}[t]
\centering
\caption{KITTI-MOS construction and geometry configurations, supplementing Table~7 in the main paper. The annotation pipeline is separate from the geometry inputs used by DynOSeg.}
\label{tab:kitti_mos_protocol}
\small
\setlength{\tabcolsep}{4pt}
\begin{tabular*}{\textwidth}{@{\extracolsep{\fill}}ll@{}}
\toprule
Item & Setting \\
\midrule
Dataset source & KITTI Raw sequences with tracklet annotations \\
Camera stream & Left color camera, \texttt{image\_02} \\
Number of sequences / frames & 38 sequences / 12,915 frames \\
Reference camera poses & KITTI GT camera poses derived from OXTS and calibration \\
Reference dense depth & FoundationStereo applied to KITTI stereo inputs \\
Alternative geometry inputs & Offline ViPE poses/depth; online DynOSAM poses + completed KITTI depth \\
Object motion source & KITTI tracklet 3D object poses + OXTS ego-motion \\
Dynamic criterion & Smoothed max horizontal object speed \(>1.5\,\mathrm{m/s}\) \\
Frame rate & \(10\) Hz \\
2D prompt generation & Projected 3D tracklet boxes on \texttt{image\_02} \\
Invalid projections & Fewer than 4 visible corners, area below 50 px, or depth below 0.1 m \\
Mask generator & SAM~3 with box and object-type text prompts \\
Final pseudo-GT mask & Union of SAM~3 masks for moving tracklets \\
\bottomrule
\end{tabular*}
\end{table*}

Few datasets simultaneously provide camera poses, dense depth maps, and moving-object masks. To evaluate DynOSeg with geometry cues from different sources, we construct the \textbf{KITTI-MOS benchmark} using KITTI tracklets and SAM~3~\cite{sam3}. KITTI camera and object poses are used to identify moving tracklets and generate pseudo-GT moving-object masks. The dense depth used by DynOSeg is supplied by the selected configuration; it is not ground-truth dense depth.

\paragraph{KITTI-MOS construction.}
\begin{figure*}[t]
  \centering
  \includegraphics[width=\linewidth]{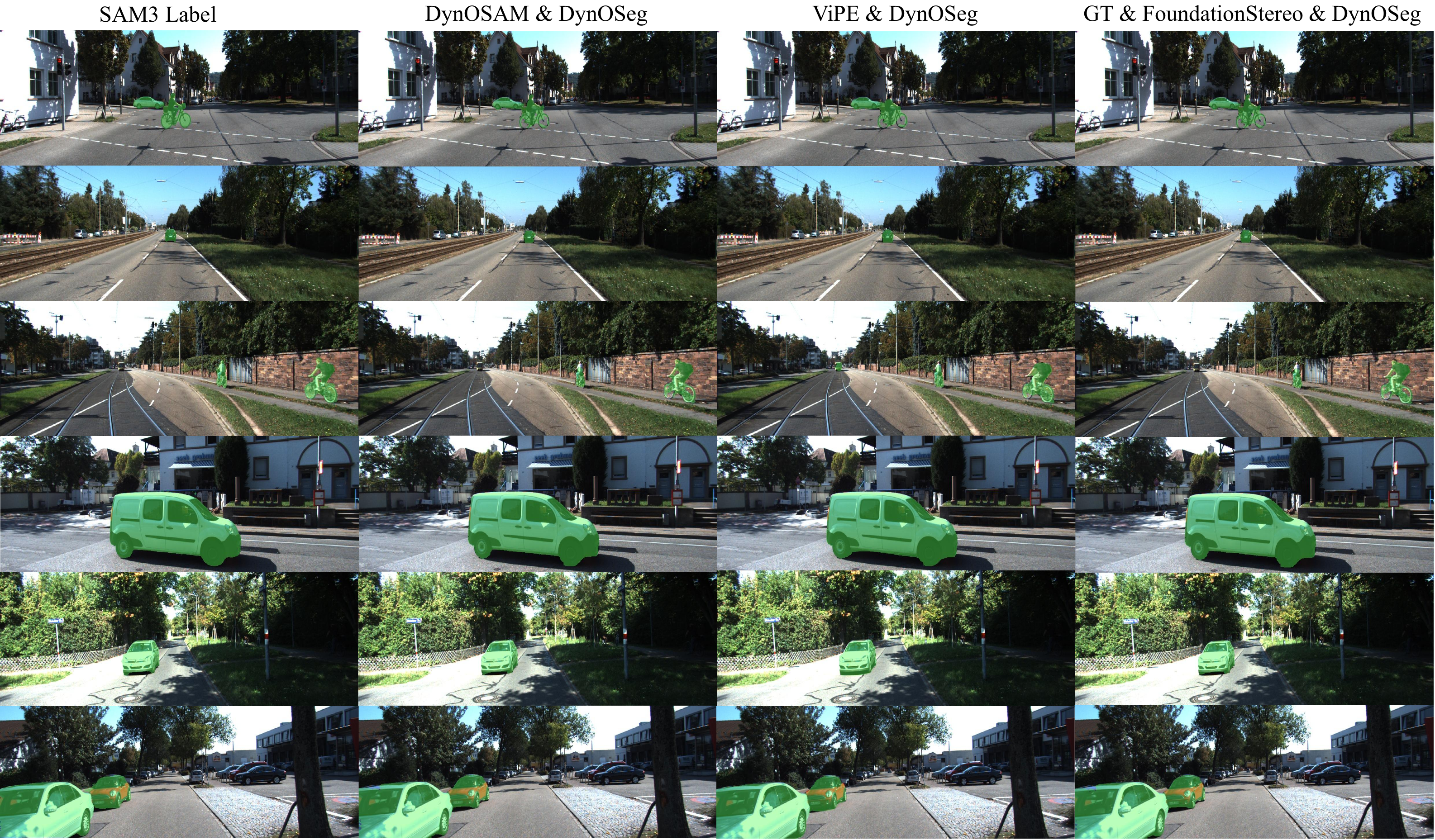}
  \caption{\textbf{Visualization results on KITTI-MOS.} The ``GT \& FoundationStereo \& DynOSeg'' configuration uses KITTI GT camera poses and FoundationStereo dense depth from KITTI stereo inputs. ``GT'' refers only to camera pose; the reference masks are SAM~3 pseudo-GT. The DynOSAM configuration uses online DynOSAM camera poses and nearest-valid completion of KITTI sparse LiDAR depth, while ViPE supplies its offline poses and depth.}
  \label{fig:sec-kitti-mos}
\end{figure*}

To support the plug-and-play geometry comparison in the main paper, we construct the KITTI-MOS benchmark from KITTI Raw sequences~\cite{Geiger2013IJRR} with available tracklet annotations. For each sequence, we use the left color camera images, KITTI calibration, OXTS ego-motion, Velodyne scans, and tracklet labels. Each tracklet provides an object category,
dimensions, and a per-frame 6-DoF object pose in the Velodyne frame. We first
recover each object's world-frame trajectory using KITTI calibration and OXTS
poses and then classify a tracklet as dynamic if its smoothed maximum horizontal
speed exceeds \(1.5\,\mathrm{m/s}\) at the KITTI Raw frame rate of \(10\) Hz.
For each visible dynamic object, we project the 3D tracklet box to the
\texttt{image\_02} plane and use the resulting 2D box, together with an
object-class text prompt, to query SAM~3~\cite{sam3}. The per-object SAM~3 masks are then
combined into a binary dynamic mask:
\begin{equation}
    M_t(\mathbf{p})
    =
    \bigvee_{i \in \mathcal{D}_t}
    \mathrm{SAM\,3}\!\left(I_t, \mathcal{B}_{i,t}, c_i\right)(\mathbf{p}),
\end{equation}
where \(I_t\) is the RGB frame, \(\mathcal{B}_{i,t}\) is the projected 2D box of
object \(i\), \(c_i\) is the text prompt derived from its KITTI object type, and
\(\mathcal{D}_t\) denotes dynamic objects visible at frame \(t\). We discard
invalid projections with fewer than four visible 3D box corners, area below
\(50\) pixels, or depth below \(0.1\) m. The final benchmark contains 38 KITTI
Raw sequences and 12,915 frames; \cref{tab:kitti_mos_protocol} summarizes the protocol.

\paragraph{Geometry-provider configurations.}
The reference configuration in the main paper combines KITTI ground-truth camera poses, obtained from OXTS ego-motion and camera calibration, with dense depth generated by FoundationStereo~\cite{wen2025stereo} from KITTI stereo inputs.
Thus, the reference does not assume access to ground-truth dense depth.
The ViPE configuration uses camera poses and depth estimated offline by ViPE~\cite{huang2025vipe}. For the online configuration, we extend the DynOSAM~\cite{morris2025dynosam} data-provider interface to accept KITTI sparse LiDAR depth and expose the depth together with the camera poses estimated online by DynOSAM. The instance masks required by DynOSAM are generated by its built-in online detector.
The same trained DynOSeg network and residual construction are used for all three configurations, without provider-specific retraining.
Table~7 in the main paper reports prediction consistency $\mathcal{C}_{\mathrm{IoU}}$ values of $89.0$, $84.0$, and $89.0$ for the KITTI reference, ViPE, and DynOSAM configurations, respectively, with an overall mean pairwise IoU of $87.3$.
These consistency scores show that the same trained DynOSeg network produces similar masks with the tested geometry providers; they do not measure geometry accuracy, robustness to arbitrary geometry errors, or agreement with human-annotated ground truth.

\paragraph{Nearest-valid depth propagation.}
KITTI Velodyne points are projected onto the left color image using the provided camera--LiDAR calibration, producing a sparse metric-depth map $D_t^{\mathrm{s}}$ at frame $t$.
If multiple projected points fall on the same pixel, we retain the smallest positive depth.
Let $\Omega_t$ denote the set of pixels with valid projected depth.
For every missing pixel $\mathbf{p}\notin\Omega_t$, the modified DynOSAM interface copies the depth of the spatially nearest valid pixel in the image plane:
\begin{equation}
    \mathbf{q}^{*}
    = \arg\min_{\mathbf{q}\in\Omega_t}
      \lVert\mathbf{p}-\mathbf{q}\rVert_2,
    \qquad
    D_t(\mathbf{p}) = D_t^{\mathrm{s}}(\mathbf{q}^{*}).
\end{equation}
The original valid LiDAR measurements are retained unchanged, i.e.,
$D_t(\mathbf{p})=D_t^{\mathrm{s}}(\mathbf{p})$ for
$\mathbf{p}\in\Omega_t$.
The completed map is stored as a double-precision metric-depth image in the format expected by DynOSAM.
This operation is deterministic, uses only the current sparse-depth map, and requires neither training nor future frames.
DynOSAM consumes this completed depth to estimate the current camera pose, and DynOSeg uses the same completed depth with the DynOSAM pose estimates.

\begin{table*}[t]
\centering
\caption{Per-sequence DynOSeg results on FBMS-59. Scores are percentages; the final row reports the unweighted sequence mean, at the same precision as Table~2 of the main paper, and the total number of annotated frames. $\dagger$ marks sequences with $\mathcal{J\&F}<70$.}
\label{tab:fbms_per_sequence}
\small
\setlength{\tabcolsep}{8pt}
\renewcommand{\arraystretch}{1.0}
\begin{tabularx}{\textwidth}{lXcccc}
\toprule
\textbf{Dataset} & \textbf{Sequence} & \multicolumn{1}{c}{\textbf{Annotated frames}} & \multicolumn{1}{c}{$\mathcal{J}\uparrow$} & \multicolumn{1}{c}{$\mathcal{F}\uparrow$} & \multicolumn{1}{c}{$\mathcal{J\&F}\uparrow$} \\
\midrule
\multirow[t]{30}{*}{FBMS-59} & camel01     &  6 & 91.28 & 99.64 & 95.46 \\
         & cars1       &  3 & 97.42 & 98.70 & 98.06 \\
\rowcolor{gray!12} & cars10$^{\dagger}$    &  4 & 69.29 & 63.50 & 66.40 \\
\rowcolor{gray!12} & cars4$^{\dagger}$     &  6 & 28.12 & 47.55 & 37.84 \\
         & cars5       &  4 & 63.25 & 82.73 & 72.99 \\
         & cats01      &  9 & 94.76 & 98.36 & 96.56 \\
         & cats03      &  6 & 88.59 & 90.87 & 89.73 \\
         & cats06      & 18 & 61.13 & 79.38 & 70.25 \\
\rowcolor{gray!12} & dogs01$^{\dagger}$    & 11 & 53.97 & 67.52 & 60.75 \\
         & dogs02      & 22 & 94.33 & 99.25 & 96.79 \\
         & farm01      & 14 & 88.25 & 81.39 & 84.82 \\
         & giraffes01  & 17 & 92.07 & 98.56 & 95.31 \\
         & goats01     & 15 & 85.64 & 94.17 & 89.90 \\
         & horses02    & 13 & 90.39 & 98.74 & 94.56 \\
         & horses04    & 41 & 91.02 & 95.10 & 93.06 \\
         & horses05    & 23 & 75.36 & 79.79 & 77.57 \\
         & lion01      &  9 & 87.47 & 89.40 & 88.44 \\
         & marple12    &  8 & 80.43 & 75.49 & 77.96 \\
         & marple2     & 11 & 86.11 & 88.68 & 87.39 \\
         & marple4     &  5 & 90.09 & 86.59 & 88.34 \\
         & marple6     & 21 & 95.05 & 89.79 & 92.42 \\
         & marple7     & 16 & 78.16 & 72.27 & 75.22 \\
         & marple9     &  8 & 89.28 & 84.20 & 86.74 \\
         & people03    & 10 & 87.99 & 83.59 & 85.79 \\
         & people1     &  5 & 92.24 & 99.41 & 95.83 \\
         & people2     &  4 & 91.94 & 94.63 & 93.29 \\
         & rabbits02   & 14 & 93.31 & 97.67 & 95.49 \\
         & rabbits03   & 12 & 91.70 & 96.39 & 94.04 \\
\rowcolor{gray!12} & rabbits04$^{\dagger}$ & 16 & 57.74 & 66.63 & 62.19 \\
         & tennis      & 16 & 90.61 & 97.44 & 94.02 \\
\midrule
\multicolumn{2}{l}{\textbf{Sequence mean / total frames}} & 367 & \textbf{82.6} & \textbf{86.6} & \textbf{84.6} \\
\bottomrule
\end{tabularx}
\end{table*}

\paragraph{Causal online interface.}
At time $t$, the interface accepts depth and camera poses for frames $t-1$ and $t$ and optical flow between the two RGB frames.
The previous frame's geometry can be buffered while the online provider produces the current estimates.
The resulting residual map is aligned with the current frame, and the segmentation network receives $(I_t,D_t,R_t)$, including the current depth $D_t$ rather than the previous depth as its depth-encoder input.
No future frame is needed when the pose/depth provider is causal and the flow is computed from this two-frame pair.
Thus, DynOSeg can be integrated into online systems using online geometry sources such as DynOSAM; this does not make the default offline ViPE configuration online.
Nor does online compatibility by itself establish real-time throughput: the full system must also account for geometry estimation, optical flow, residual construction, data transfer, and segmentation.

\paragraph{Geometry-induced residual.}
For all geometry sources, DynOSeg uses the same residual construction.  Given
source frame \(t\), adjacent target frame \(q\), camera-to-world poses
\(T_{w \leftarrow c_t}\) and \(T_{w \leftarrow c_q}\), depth \(D_t\), optical
flow \(\mathbf{f}_{t \rightarrow q}\), and intrinsics \(K\), we compute
\begin{equation}
    T_{c_q \leftarrow c_t}
    =
    T_{w \leftarrow c_q}^{-1} T_{w \leftarrow c_t},
\end{equation}
\begin{equation}
    \hat{\mathbf{f}}_{t \rightarrow q}(\mathbf{p})
    =
    \pi\!\left(
        K\!\left[
        \mathbf{R}_{c_q \leftarrow c_t}
        D_t(\mathbf{p}) K^{-1} \tilde{\mathbf{p}}
        + \mathbf{v}_{c_q \leftarrow c_t}
        \right]
    \right)
    -
    \mathbf{p},
\end{equation}
where $\mathbf{R}_{c_q\leftarrow c_t}$ and $\mathbf{v}_{c_q\leftarrow c_t}$ are the rotation and translation of $T_{c_q\leftarrow c_t}$, $\tilde{\mathbf{p}}$ is a homogeneous pixel coordinate, and $\pi$ performs perspective division.
Depth and translation must use a common scale.
The residual magnitude is
\begin{equation}
    R_t(\mathbf{p})
    =
    \left\|
    \mathbf{f}_{t \rightarrow q}(\mathbf{p})
    -
    \hat{\mathbf{f}}_{t \rightarrow q}(\mathbf{p})
    \right\|_2 .
\end{equation}

For online inference, $q=t-1$ gives a current-grid residual using current-to-previous optical flow, following Section~3.1 of the main paper.
This residual map is used as the motion cue for DynOSeg. Importantly, only the
geometry source changes across the KITTI-MOS evaluation in the main paper; RGB frames, optical
flow, and evaluation masks remain fixed.

\paragraph{Annotation and geometry limitations.}
KITTI-MOS evaluates binary moving regions induced by the tracklet and SAM~3 annotation pipeline, not manually verified instance masks.
Tracklet coverage and prompting errors can affect both apparent false positives and false negatives.
The geometry comparison holds these pseudo-GT masks fixed; it does not measure annotation quality.
Depth, pose, or flow errors can also produce nonzero residuals in static regions.
Consequently, high prediction consistency across the three providers is not evidence that their geometry estimates are equally accurate.

\section{FBMS-59 Per-Sequence Results}
\label{appendix:fbms_per_sequence}

For completeness, \cref{tab:fbms_per_sequence} reports the DynOSeg result for every evaluated FBMS-59 sequence together with its number of annotated frames.
The per-sequence metrics are expressed as percentages and rounded to two decimals.
The final row is the unweighted mean over the 30 sequence-level scores, shown to one decimal to exactly match the FBMS-59 entry in Table~2 of the main paper; annotated frames are summed across sequences.
A dagger marks the four sequences with $\mathcal{J\&F}<70$.

\begin{figure*}[t]
\centering

\includegraphics[width=\linewidth]{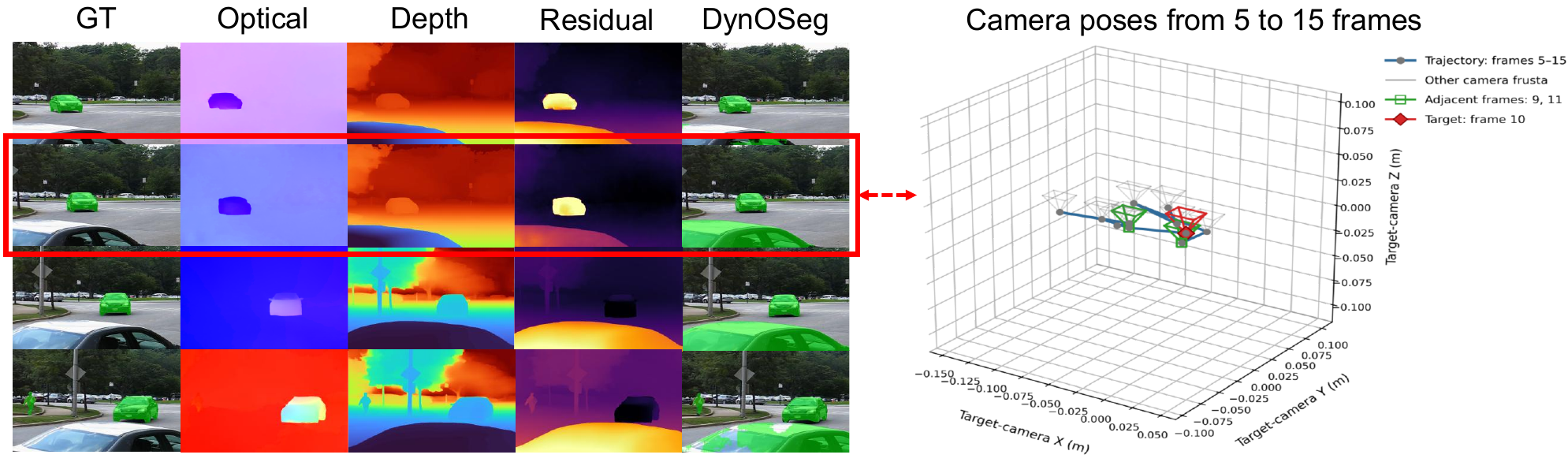}

\caption{Failure analysis on the FBMS-59 \texttt{cars4} sequence. The left panel shows four representative examples. From left to right, each example contains the ground-truth foreground overlay, optical flow, metric depth, motion residual, and DynOSeg prediction. The red rectangle marks target frame 10 and links it to the camera-pose visualization on the right. The trajectory covers frames 5--15, with frame 10 shown in red and adjacent frames 9 and 11 shown in green. Metric depth uses the Turbo colormap, where cooler colors indicate smaller depth and warmer colors indicate larger depth. Motion residuals use Inferno, where darker colors indicate better agreement with rigid camera-induced motion and brighter colors indicate larger deviations. For optical flow, hue indicates direction and saturation indicates magnitude. All maps are normalized independently for each frame, so colors should not be compared across examples.}

\label{fig:fbms_cars4_failure}
\end{figure*}

\paragraph{Failure analysis on \texttt{cars4}.}
The \texttt{cars4} sequence has the lowest $\mathcal{J\&F}$ score in \cref{tab:fbms_per_sequence}.
As illustrated in \cref{fig:fbms_cars4_failure}, a plausible contributor is local camera-pose error around the highlighted target frame.
An inaccurate pose perturbs the rigid camera-induced flow synthesized from pose and depth, so static pixels can receive spurious residual responses.
DynOSeg may then interpret these residual artifacts as independent motion, which is consistent with the false-positive foreground regions near the bottom of the predictions in the later examples.
This visualization supports pose-induced residual error as a failure hypothesis rather than establishing it as the sole cause. Depth and optical-flow errors can produce similar effects.

\section{Runtime and Resolution Scaling}
\label{appendix:supp_runtime}

\paragraph{Timing definitions.}
All runtimes use one NVIDIA RTX 4090 with batch size $1$.
First, Table~8 in the main paper and~\cref{tab:supp_runtime} isolate how the main components scale with input resolution.
DynOSeg wall-clock timing starts after the depth and residual maps are available and includes DINOv2 feature extraction, data transfer, feature fusion, decoding, and mask post-processing.
It therefore excludes geometry estimation, RAFT, residual-map construction, and disk I/O.
The separate DynOSeg GPU column measures DINOv2 and segmentation computation only; it excludes data transfer and post-processing.
RAFT uses the FlyingThings3D checkpoint with $20$ iterations, and the residual column measures camera-induced flow and residual-map construction once depth, poses, and optical flow are available.
Because the wall-clock and GPU measurements were collected independently, their difference does not isolate transfer or post-processing overhead, and the columns in~\cref{tab:supp_runtime} must not be added to claim measured end-to-end latency.

\paragraph{Causal online and offline pipelines.}
DynOSeg requires only the current and previous frames, their depth and camera poses, and the optical flow between them, as detailed in~\cref{sec:kitti-mos}.
The previous frame can be buffered, so no future frame is required when the geometry provider is causal.
The evaluated online route uses the modified DynOSAM interface described above, followed by RAFT, DynOSAM pose estimation, residual construction, and DynOSeg inference.
The nearest-valid sparse-to-dense preprocessing is outside the scope of this timing analysis and is excluded from the reported online totals.
The offline route uses ViPE to estimate camera poses and depth with sequence-level processing, followed by RAFT, residual construction, and DynOSeg.
The GT-pose plus FoundationStereo route is a controlled reference rather than a deployable online system because it uses KITTI ground-truth camera poses.

\begin{center}
\captionsetup{hypcap=false}
\captionof{table}{Runtime at five resolutions, in ms/frame. These are independently measured component timings, not additive stages of an end-to-end benchmark. Values match Table~8 in the main paper.}
\label{tab:supp_runtime}
\footnotesize
\setlength{\tabcolsep}{3pt}
\begin{tabular*}{\columnwidth}{@{\extracolsep{\fill}}lrrrr@{}}
\toprule
Resolution & \multicolumn{2}{c}{DynOSeg} & RAFT & Residual \\
$(W\times H)$ & Wall-clock & GPU & GPU & GPU \\
\midrule
$518\times518$   & 20.13  & 19.00  & 25.93  & 0.43 \\
$630\times630$   & 32.95  & 32.28  & 31.99  & 0.44 \\
$728\times728$   & 46.93  & 47.15  & 44.89  & 0.46 \\
$1344\times756$  & 102.43 & 106.20 & 84.57  & 0.58 \\
$1792\times1008$ & 252.69 & 270.65 & 173.47 & 1.02 \\
\bottomrule
\end{tabular*}
\end{center}

\begin{table*}[t]
\centering
\caption{\textbf{Stage-wise KITTI-MOS runtime on an NVIDIA RTX 4090.}
Panel (a) lists the accounted upstream costs that do not change with the DynOSeg input size; panel (b) reports DynOSeg latency and the resulting sequential total/FPS for each route.
FS denotes FoundationStereo.}
\label{tab:kitti_mos_full_pipeline_runtime}
\footnotesize
\setlength{\tabcolsep}{6pt}
\renewcommand{\arraystretch}{1.05}

\textbf{(a) Resolution-invariant upstream stages (ms/frame)}

\vspace{2pt}
\begin{tabular*}{0.78\textwidth}{@{\extracolsep{\fill}}lrrrr@{}}
\toprule
Route & Geometry/depth & RAFT & DynOSAM & Residual \\
\midrule
Controlled (GT pose + FS) & 180.71 & 27.81 & --    & 0.48 \\
Offline (ViPE)            & 529.46 & 27.81 & --    & 0.48 \\
Online (DynOSAM)          & Excluded & 27.81 & 69.40 & 0.48 \\
\bottomrule
\end{tabular*}

\vspace{7pt}
\textbf{(b) Resolution-dependent DynOSeg latency and accounted sequential estimates}

\vspace{2pt}
\begin{tabular*}{0.92\textwidth}{@{\extracolsep{\fill}}lrrrr@{}}
\toprule
\multirow{2}{*}{DynOSeg input}
& \multirow{2}{*}{DynOSeg (ms)}
& \multicolumn{3}{c}{Sequential total (ms/frame) / FPS} \\
\cmidrule(lr){3-5}
& & Controlled & Offline & Online \\
\midrule
$196\times196$   & 4.98   & 213.99 / 4.67 & 562.73 / 1.78 & 102.67 / 9.74 \\
$392\times392$   & 9.38   & 218.39 / 4.58 & 567.14 / 1.76 & 107.07 / 9.34 \\
$588\times588$   & 13.75  & 222.76 / 4.49 & 571.51 / 1.75 & 111.44 / 8.97 \\
$784\times784$   & 20.63  & 229.63 / 4.35 & 578.38 / 1.73 & 118.32 / 8.45 \\
$980\times980$   & 32.92  & 241.93 / 4.13 & 590.68 / 1.69 & 130.61 / 7.66 \\
$1176\times1176$ & 54.23  & 263.24 / 3.80 & 611.98 / 1.63 & 151.92 / 6.58 \\
$1372\times1372$ & 84.81  & 293.81 / 3.40 & 642.56 / 1.56 & 182.50 / 5.48 \\
$1568\times1568$ & 126.00 & 335.01 / 2.98 & 683.76 / 1.46 & 223.69 / 4.47 \\
$1764\times1764$ & 187.50 & 396.50 / 2.52 & 745.25 / 1.34 & 285.19 / 3.51 \\
$1932\times1932$ & 256.17 & 465.17 / 2.15 & 813.92 / 1.23 & 353.86 / 2.83 \\
\bottomrule
\end{tabular*}

\begin{minipage}{0.94\textwidth}
\footnotesize
\textit{Protocol.}
Upstream stages run at the native KITTI resolution ($1242\times375$); square DynOSeg inputs use multiples of 14.
FoundationStereo is averaged over 50 runs, and RAFT, residual construction, and DynOSeg over 30 runs with batch size 1.
Totals sum independently averaged stages without pipeline overlap and exclude disk I/O; the online totals also exclude nearest-valid depth propagation.
\end{minipage}
\end{table*}
\paragraph{Sequential runtime accounting.}
\Cref{tab:kitti_mos_full_pipeline_runtime} summarizes stage-wise latency accounting on KITTI-MOS without repeating the resolution-invariant upstream costs in every row.
Panel (a) lists the accounted resolution-invariant stages for each route, while panel (b) varies only the square DynOSeg input resolution and reports the resulting sequential totals.
The reported total is the sequential sum of independently averaged stage times without pipeline overlap; it is not a directly measured end-to-end wall-clock latency.
FPS is the reciprocal of this sequential estimate.

\paragraph{Component scaling.}
The tested dimensions in~\cref{tab:supp_runtime} are multiples of DINOv2's patch size $14$; the two rectangular settings use a $16{:}9$ aspect ratio.
With precomputed depth and residual maps, DynOSeg wall-clock latency increases from $20.13$ ms/frame at $518\times518$ to $252.69$ ms/frame at $1792\times1008$.
The former corresponds to approximately $49.7$ FPS for the DynOSeg module alone, whereas the latter corresponds to approximately $4.0$ FPS.
Across the same settings, RAFT requires $25.93$--$173.47$ ms/frame, while residual construction requires only $0.43$--$1.02$ ms/frame.
Thus, once depth, poses, and optical flow are available, converting them into the residual cue is inexpensive relative to optical-flow estimation and segmentation.

\paragraph{Accounted pipeline cost and bottlenecks.}
At a $196\times196$ DynOSeg input, the controlled reference, offline ViPE, and online DynOSAM routes have accounted sequential estimates of $213.99$, $562.73$, and $102.67$ ms/frame, respectively, corresponding to $4.67$, $1.78$, and $9.74$ FPS.
At $1932\times1932$, these estimates increase to $465.17$, $813.92$, and $353.86$ ms/frame ($2.15$, $1.23$, and $2.83$ FPS).
At the smallest DynOSeg resolution, ViPE alone requires $529.46$ ms/frame in the offline route, while the accounted online route includes $27.81$ ms for RAFT and $69.40$ ms for the DynOSAM stage, compared with $4.98$ ms for DynOSeg.
At the largest resolution, DynOSeg grows to $256.17$ ms/frame and becomes a substantial part of the total, although ViPE remains the largest stage in the offline route.
Residual construction remains $0.48$ ms/frame at native KITTI resolution and contributes less than $0.5\%$ of every total in~\cref{tab:kitti_mos_full_pipeline_runtime}.

\paragraph{Online runtime analysis.}
Here, \emph{online} means causal: the DynOSAM configuration can produce the current prediction without observing future frames.
It does not mean that the current implementation runs at the input frame rate.
The fastest accounted online sequential estimate is $102.67$ ms/frame ($9.74$ FPS), which already exceeds the $100$ ms/frame budget of KITTI's $10$ Hz stream before the excluded sparse-to-dense preprocessing is considered.
Accordingly, our results demonstrate compatibility with online geometry inputs, not real-time system throughput.
Achieving real-time operation would require a directly optimized and measured streaming implementation, potentially with faster geometry and flow providers, lower DynOSeg resolution, and pipeline overlap.

\end{document}